\pdfoutput=1

\documentclass[11pt]{article}

\usepackage{acl}

\usepackage{times}
\usepackage{latexsym}
\usepackage{times,latexsym}
\usepackage{url}
\usepackage[T1]{fontenc}
\usepackage{booktabs}       %
\usepackage{amsfonts}       %
\usepackage{nicefrac}       %
\usepackage{microtype}      %
\usepackage{xcolor}         %
\usepackage{graphicx}
\usepackage{bm}             %
\usepackage{longtable}
\usepackage[toc, page]{appendix}
\usepackage{hyperref}
\usepackage{subcaption}   

\usepackage[T1]{fontenc}

\usepackage[utf8]{inputenc}

\usepackage{microtype}

\title{Characterizing Verbatim Short-Term Memory in Neural Language Models}

\author{
 Kristijan Armeni \\
 Johns Hopkins University\\
 \href{mailto:karmeni1@jhu.edu}{\nolinkurl{karmeni1@jhu.edu}}
  \And
  Christopher Honey \\
  Johns Hopkins University \\
  \href{mailto:chris.honey@jhu.edu}{\nolinkurl{chris.honey@jhu.edu}}
  \And 
  Tal Linzen \\
  New York University \\
  \href{mailto:linzen@nyu.edu}{\nolinkurl{linzen@nyu.edu}}
}

\begin{document}
\maketitle
\begin{abstract}
When a language model is trained to predict natural language sequences, its prediction at each moment depends on a representation of prior context. What kind of information about the prior context can language models retrieve? We tested whether language models could retrieve the exact words that occurred previously in a text. In our paradigm, language models (transformers and an LSTM) processed English text in which a list of nouns occurred twice. We operationalized retrieval as the reduction in surprisal from the first to the second list. We found that the transformers retrieved both the identity and ordering of nouns from the first list. Further, the transformers' retrieval was markedly enhanced when they were trained on a larger corpus and with greater model depth. Lastly, their ability to index prior tokens was dependent on learned attention patterns. In contrast, the LSTM exhibited less precise retrieval, which was limited to list-initial tokens and to short intervening texts. The LSTM's retrieval was not sensitive to the order of nouns and it improved when the list was semantically coherent. We conclude that transformers implemented something akin to a working memory system that could flexibly retrieve individual token representations across arbitrary delays; conversely, the LSTM maintained a coarser and more rapidly-decaying semantic gist of prior tokens, weighted toward the earliest items.
\end{abstract}

\section{Introduction}
Language models (LMs) are computational systems trained to predict upcoming tokens based on past context.
To perform this task well, they must construct a coherent representation of the text, which requires establishing relationships between words that occur at non-adjacent time points.

\begin{figure}[ht]
\centering
\includegraphics[width=1\linewidth]{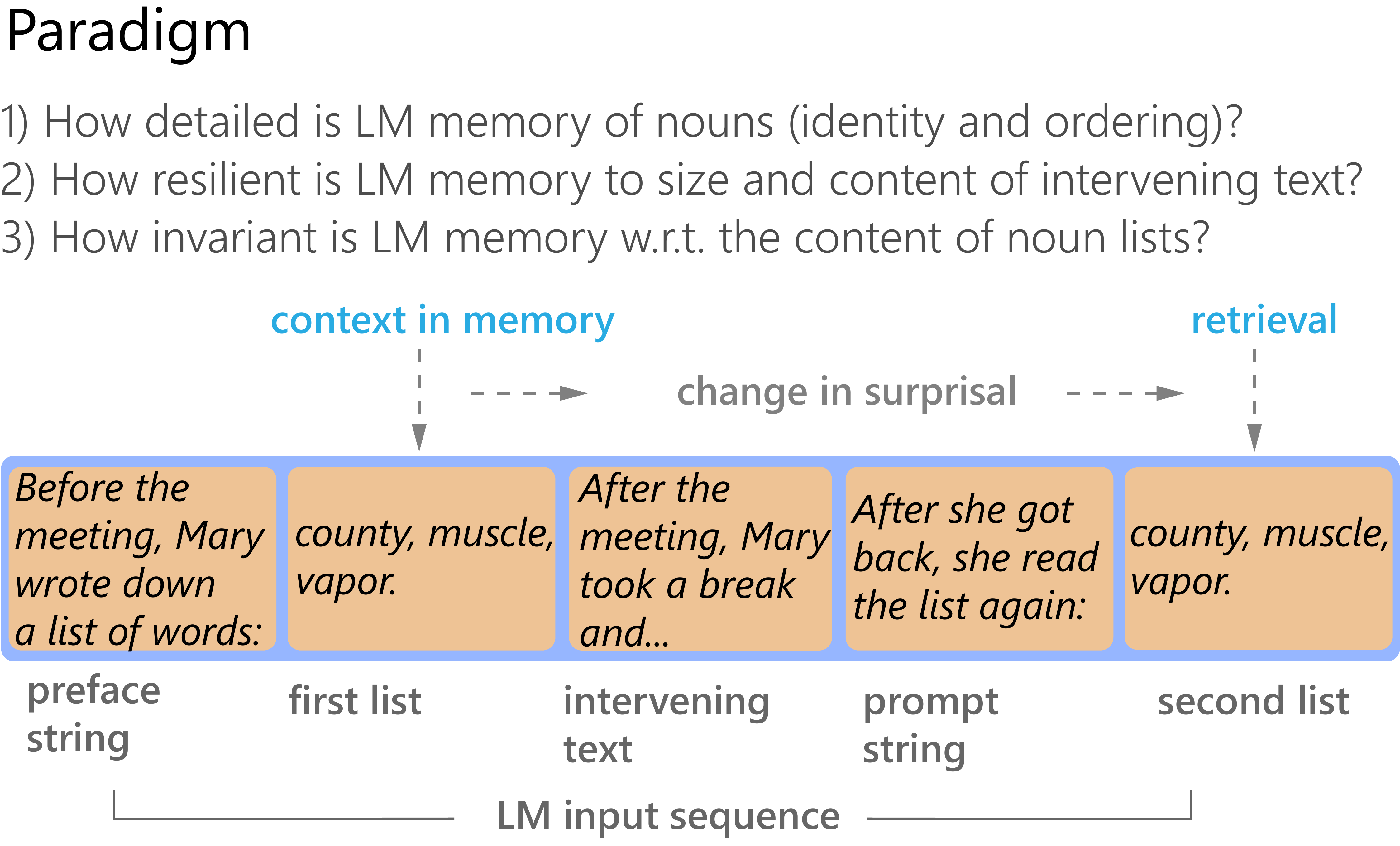}
\caption{Characterizing verbatim memory retrieval in neural language models. In our paradigm, language models processed English text in which a list of nouns occurred twice.
We operationalized retrieval as the reduction in surprisal from the first to the second list presentation.
We measured retrieval while varying: a) set size, b) the structure of the second list, c) the length of the intervening text, and d) the content and structure of the intervening text.}
\label{fig:design}
\end{figure}

Despite their simple learning objective, LMs based on contemporary artificial neural network architectures perform well in contexts that require maintenance and retrieval of dependencies spanning multiple words.
For example, LMs learn to correctly match the grammatical number of the subject and a corresponding verb across intervening words; for example, they prefer the correct \emph{The \textbf{girls} standing at the desk \textbf{are} tall}, to the incorrect \emph{The \textbf{girls} standing at the desk \textbf{is} tall} \citep{linzen_assessing_2016, marvin_targeted_2018, gulordava_colorless_2018, futrell_rnns_2018}.
The ability to maintain context across multiple words is likely to be a central factor explaining the success of these models, potentially following fine-tuning, in natural language processing tasks \citep{devlin_bert_2019, brown_language_2020}.

The work discussed above has shown that LMs extract linguistically meaningful signals and that, over the course of learning, they develop a short-term memory capacity: the ability to store and access recent past context for processing, possibly akin to the working memory systems thought to enable flexible human cognitive capacities \citep{baddeley_working_2003}. What is the nature of the memory processes that LMs learn? Are these memory processes able to access individual tokens from the recent past \textit{verbatim}, or is the memory system more implicit, so that only an aggregate \textit{gist} of the prior context is available to subsequent processing? 

Here, we introduce a paradigm (Fig.~\ref{fig:design}), inspired by benchmark tasks for models of human short-term memory \citep{oberauer_benchmarks_2018}, for characterizing short-term memory abilities of LMs.
We apply it to two particular neural LM architectures that possess the architectural ingredients to hold past items in memory: attention-based transformers \citep{vaswani_attention_2017} and long short-term memory networks \citep[LSTM]{hochreiter_long_1997}.
Whereas LSTMs incorporate the past by reusing the results of processing from previous time steps through dedicated memory cells, transformers use the internal representations of each of the previous tokens as input.
These architectural ingredients alone, however, are not sufficient for a model to have memory.
We hypothesize that whether or not the model puts this memory capacity to \textit{use} depends on whether the training task (next word prediction) requires it --- the parameters controlling the activation of context representations and subsequent retrieval computations are in both cases \textit{learned}.

Our goal is to determine whether and when the LMs we study maintain and retrieve verbatim representations of individual prior tokens. 
First, we measure the \textit{detail} of the context representation: does the LM maintain a verbatim representation of all prior tokens and their order, or does it instead combine multiple prior tokens into a summary representation, like a semantic gist? 
Second, we consider the \textit{resilience} of the memory to interference: after how many intervening tokens do the representation of prior context become inaccessible?
Third, we consider the \textit{content-invariance} of the context representations: does the resilience of prior context depend on semantic coherence of the prior information, or can arbitrary and unrelated information sequences be retrieved?

\section{Related Work}

Previous studies examined how properties of linguistic context influenced next-word prediction accuracy in transformer and LSTM LMs trained on text in English.
\citet{khandelwal_sharp_2018} showed that LSTM LMs use a window of approximately 200 tokens of past context and word order information of the past 50 words, in the service of predicting the next token in natural language sequences. \citet{subramanian_multi-scale_2020} applied a similar analysis to a transformer LM and showed that LM loss on test-set sequences was not sensitive to context perturbations beyond 50 tokens.
\citet{oconnor_what_2021} investigated whether fine-grained lexical and sentential features of context are used for next-word prediction in transformer LMs.
They showed that transformers rely predominantly on local word co-occurrence statistics (e.g. trigram ordering) and the presence of open class parts of speech (e.g. nouns), and less on the global structure of context (e.g. sentence ordering) and the presence of closed class parts of speech (e.g. function words).
In contrast with these studies, which focused on how specific features of past context affect LM performance on novel input at test time, our paradigm tests for the ability of LMs to retrieve nouns that are exactly repeated from prior context.

In a separate line of work bearing on memory maintenance in LSTMs, \citet{lakretz_emergence_2019, lakretz_mechanisms_2021} studied an LSTM's capacity to track subject-verb agreement dependencies. 
They showed that LSTM LMs relied on a small number of hidden units and the gating mechanisms that control memory contents. 
Here, we are similarly concerned with memory characteristics that support LM performance, but — akin to behavioral tests in cognitive science — we infer the \textit{functional properties} of LM memory by manipulating properties of repeated noun lists and observing the effects these manipulations have on the behavior (surprisal) of the LM rather than on its internal representation.

A third related area of research proposes \textit{architectural} innovations that augment RNNs and LSTMs with dedicated memory components \citep[e.g.][]{weston_memory_2015, yogatama_memory_2018} or improve the handling of context and memory in transformers \citep[see][for review]{tay_efficient_2020}.
Here, we are not concerned with improving architectures, but with developing a paradigm that allows us to study how LMs put to use their memory systems, whether those are implicit or explicit.

\section{Methods}\label{sec:methods}

\subsection{Paradigm: Lists of Nouns in Context}\label{sec:paradigms}

Noun lists were embedded in brief vignettes (Figure \ref{fig:design}, A and B).
Each vignette opened with a \emph{preface string} (e.g. ``Before the meeting, Mary wrote down the following list of words:'').
This string was followed by a list of nouns (the \emph{first list}), which were separated by commas; the list-final noun was followed by a full stop (e.g. ``county, muscle, vapor.'').
The first list was followed by an \emph{intervening text}, which continued the narrative established by the preface string (``After the meeting, she took a break and had a cup of coffee.'').
The intervening text was followed by a short \emph{prompt} string (e.g. ``After she got back, she read the list again:'') after which another list of nouns, either identical to the first list or different from it, was presented (we refer to this list as the \emph{second list}).
The full vignettes are provided in Section \ref{sec:vignettes} of the Appendix.

\subsection{Semantic Coherence of Noun Lists}\label{sec:noun_lists}

We used two types of word lists: arbitrary and semantically coherent.
Arbitrary word lists (e.g. ``device, singer, picture'') were composed of randomly sampled nouns from the Toronto word pool.\footnote{\url{http://memory.psych.upenn.edu/files/wordpools/nouns.txt}}
Semantically coherent word lists were sampled from the categorized noun word pool,\footnote{\url{http://memory.psych.upenn.edu/files/wordpools/catwpool.txt}} 
which contains 32 lists, each of which contains 32 semantically related nouns (e.g. ``robin, sparrow, heron, ...'').
All noun lists used in experiments are reported in Tables \ref{tab:arbitrary_nouns} and \ref{tab:semantic_nouns} of the Appendix.

After ensuring there were at least 10 valid, in-vocabulary nouns per semantic set (as this was the maximal list length we considered), we were able to construct $23$ nouns lists.
Finally, to reduce the variance attributable to tokens occurring in specific positions, we generated 10 ``folds'' of each list by circularly shifting the tokens in the first list 10 times.
In this way, each noun in each list was tested in all possible ordinal positions.
This procedure resulted in a total of $23 \times 10 = 230$ noun lists.

\subsection{Language Models}\label{sec:models}

\paragraph{LSTM} We used an adaptive weight-dropped (AWD) LSTM released by \citet{merity_regularizing_2018}\footnote{Our code is available at: \url{https://github.com/KristijanArmeni/verbatim-memory-in-NLMs}. Our experiment data are available at: \url{https://doi.org/10.17605/OSF.IO/5GY7X}}, which had 3 hidden layers with 400-dimensional input embeddings, 1840-dimensional hidden states, and a vocabulary size of 267,735.
The model contained 182.3 million trainable parameters. 
It was trained on the Wikitext-103 corpus \citep[][]{merity_pointer_2016} and achieved a test-set perplexity of 41.8. Full training hyperparameters are reported in Section \ref{sec:lstm_training_details} of the Appendix.

\paragraph{Transformer} We trained a transformer LM on the Wikitext-103 benchmark.
We retrained the BPE tokenizer on the concatenated Wikitext-103 training, evaluation, and test sets and set. 
The vocabulary had 28,439 entries. 
We trained both the 12-layer GPT-2 architecture (known as ``GPT-2 small'', 107.7 million trainable parameters) and, as a point of comparison, smaller, 1-, 3-, and 6-layer transformers (29.7, 43.9, and 65.2 million trainable parameters, respectively). 
The context window was set to 1024 tokens and embedding dimension was kept at 768 across the architectures.
The perplexities for the 12-, 6-, 3- and 1-layer models on the Wikitext-103 test set were 40.6, 51.5, 60.1, and 95.1, respectively. The full transformer training details are reported in Section \ref{sec:transformer_training_details} of the Appendix.

We also evaluated the transformer LM pretrained by \citet{radford_language_2019}, accessed through the Hugging Face Transformers library \cite{wolf_transformers_2020}. We refer to this model simply as \mbox{GPT-2}.
It was trained on the WebText corpus, which consists of approximately 8 million online documents. 
We used the GPT-2-small checkpoint which has 12 attention layers and 768-dimensional embedding layer.
The model contains 124 million parameters and has a vocabulary of 50,257 entries.
We used the maximum context size of 1024 tokens.

\subsection{Surprisal}\label{sec:surprisal}

For each token $w_t$ in our sequence, we computed the negative log likelihood (surprisal): $\texttt{surprisal}(w_t) = -\log_{2} P(w_t | w_{1}, \ldots, w_{t-1}) \label{eq:surprisal}$.
In cases when the transformer byte-pair encoding tokenizer split a noun into multiple tokens---e.g. ``sparrow'' might be split into ``sp'' and ``arrow''---we summed the surprisals of the resulting tokens.

\paragraph{Quantifying retrieval: repeat surprisal} 
To quantify how the memory trace of the first list affected the model's expectations on the second list, we measured the ratio between the surprisal on the second list and the surprisal on the first list: $\texttt{repeat surprisal} = \frac{\bar{s}(L_2)}{\bar{s}(L_1)}\times 100 \label{eq:relative_surprisal}$, where $\bar{s}(L_1)$ refers to mean surprisal across non-initial nouns in the first list and $\bar{s}(L_2)$ to mean surprisal across all non-initial nouns in the second list.
We take a \textit{reduction} in surprisal on second lists to indicate the extent to which an LM has retrieved tokens from the first list.

\begin{figure*}[ht]
\centering
\includegraphics[width=0.7\linewidth]{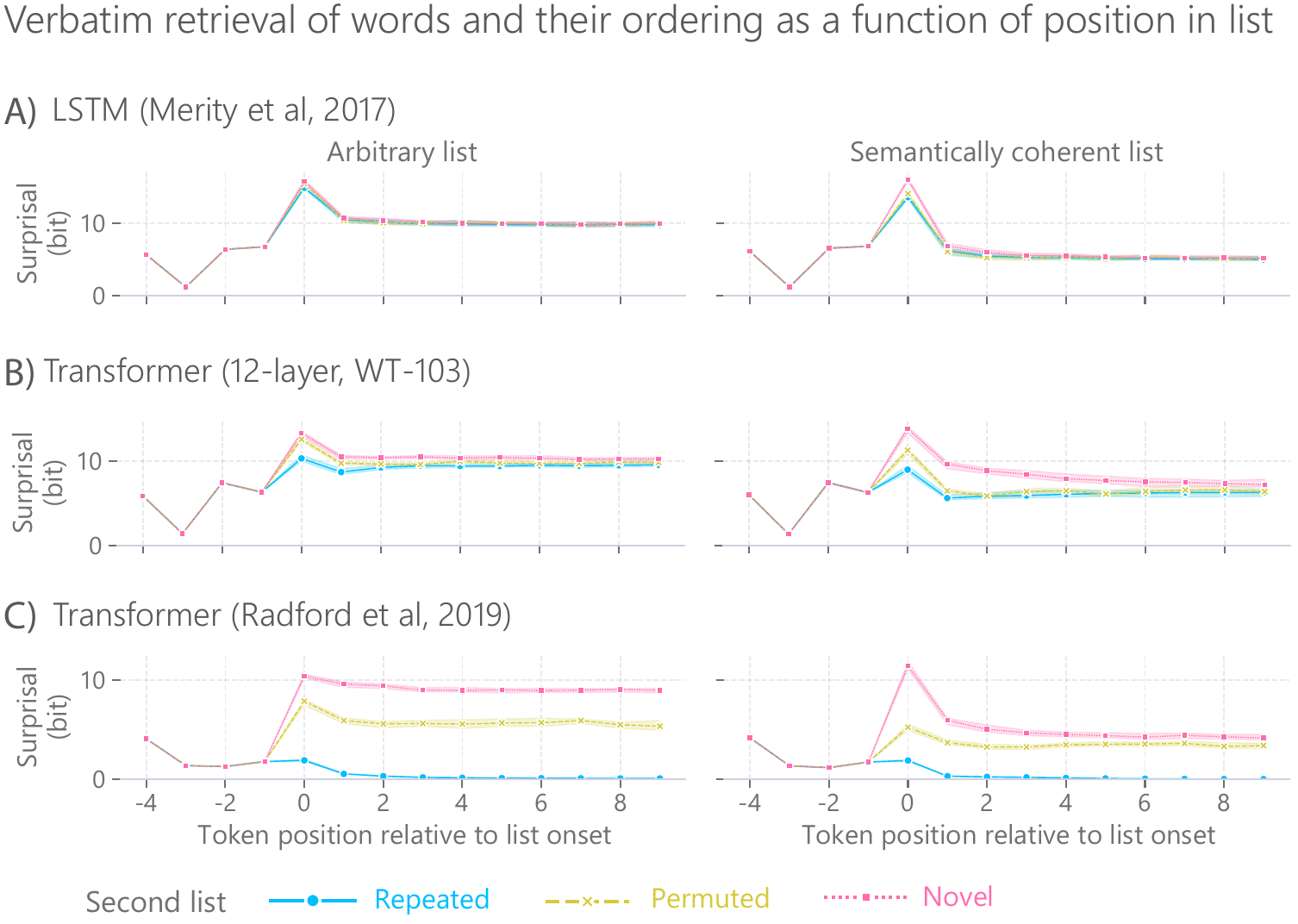}
\caption{Median surprisal (over $N^{list} = 230$) broken down per token position in second lists of arbitrary nouns and semantically coherent nouns. 
Negative values on x-axis represent 4 tokens of prompt string that introduced the second list: ``(she) read the list again''.
The 0-index marks the first noun in the list.
Line style and hue denote manipulation of the second list relative to the first list.
Error bands denote 95\% confidence interval around the median (bootstrap estimate).} 
\label{fig:per-token}
\end{figure*}

\section{Transformer Results} \label{experiments}

We first describe the results of our experiments with the two largest transformer models, the off-the-shelf GPT-2 and the 12-layer transformer we trained; LSTM results are discussed in Section~\ref{sec:lstm}, and results with smaller transformers are discussed towards the end of this section.

\paragraph{The transformers retrieved prior nouns and their order; this capacity improved when the model was trained on a larger corpus.} \label{ex:word_order}
We tested whether the transformers could retrieve the identity and order of 10-token noun lists (arbitrary or semantically coherent).
To this end, we constructed vignettes in which the second list was either (a) identical to the first list, (b) a permutation of the first list, or (c) a list of novel nouns not present in the first list.\footnote{Novel nouns in the string were introduced by randomly selecting a list of nouns from one the 22 remaining lists in the noun pool. In semantically coherent lists, novel nouns were drawn from a different semantic category than the nouns in the first list.}
We then measured retrieval as reduction in surprisal from first to second list.

When the two transformers were presented with second lists that were repeated version of the first ones (blue in Fig.~\ref{fig:per-token}, B and C), token-by-token surprisal decreased compared to novel tokens, suggesting that the transformers were able to access verbatim representations of past nouns from context.
When the second list was a permutation of the first one, surprisal was higher compared to when it was repeated, indicating that the transformers expected the nouns to be ordered as in the first list.
Training set size played an important role in supporting verbatim recall: surprisal differences were considerably smaller for the transformer trained on the Wikitext-103 corpus (Fig.~\ref{fig:per-token},~B) compared to GPT-2 (Fig.~\ref{fig:per-token},~C).

\begin{figure*}
\centering
\includegraphics[width=0.7\linewidth]{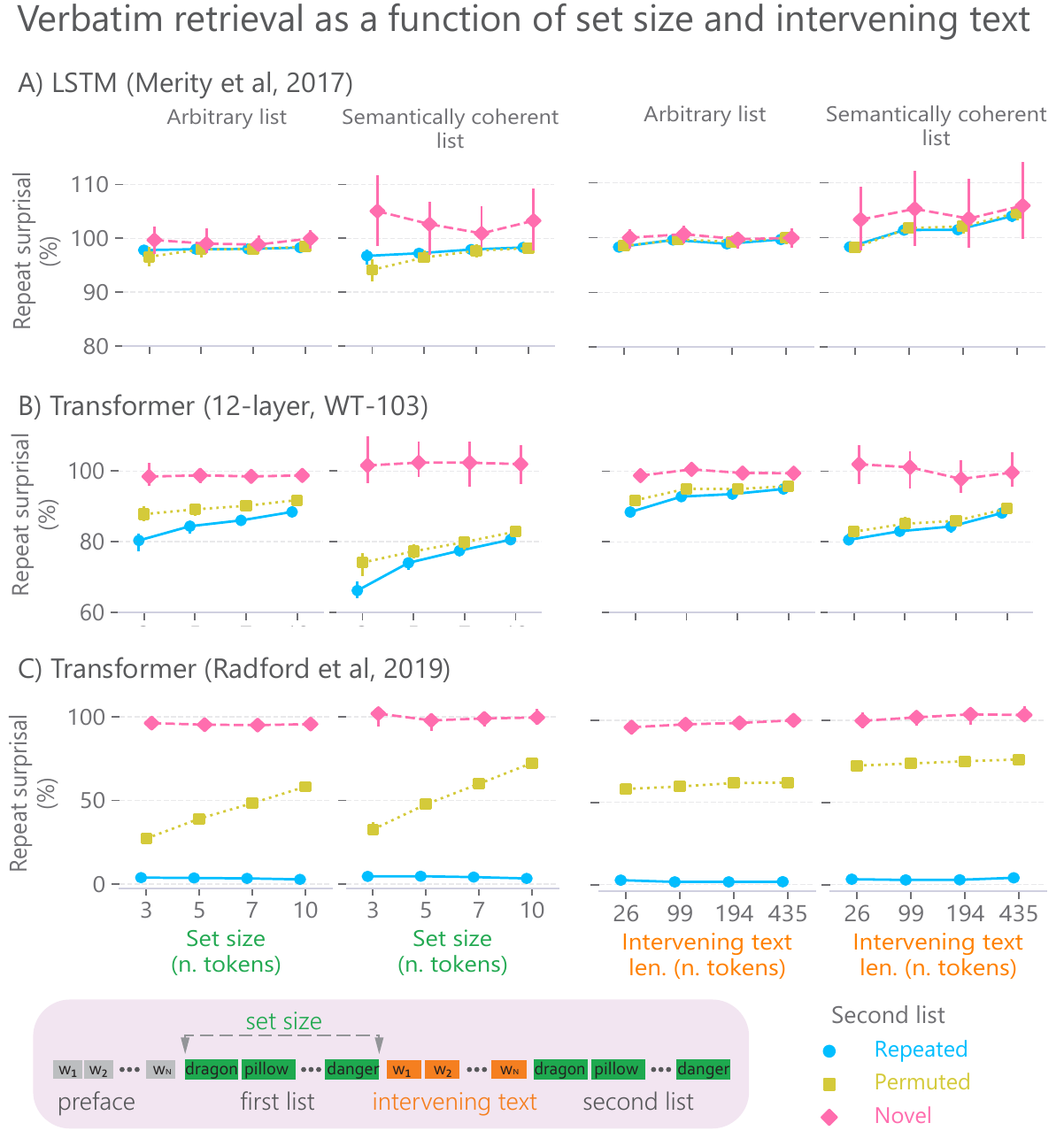}
\caption{Verbatim token retrieval for varying number of tokens being retrieved (left) and the length of the intervening text (right).
Reported is proportion of list-averaged surprisal on second relative to first list of nouns.
Points show group median (over $N^{list}$ = 230). 
Error bars denote 95\% confidence interval around the median (bootstrap estimate).
For set size manipulation, intervening text is fixed at 26 tokens. For intervening text manipulation, set size is fixed at 10 tokens.}
\label{fig:set-size}
\end{figure*}

In order to contextualize the magnitude of these retrieval effects, we computed the relative surprisal across all tokens in lists except the first one (Fig.~\ref{fig:set-size}).
When the first and second lists were identical (e.g. with $N=10$ arbitrary nouns), the Wikitext-103 transformer's median relative surprisal was at $88\%$ of the first list, compared to $92\%$ for the permuted lists, and $99\%$ for the novel lists.
In GPT-2, repeat surprisal was only $2\%$ of the first list, much lower than the $58\%$ for the permuted lists, and $96\%$ of the novel list.

Retrieval in GPT-2 was robust to the exact phrasing of the text that introduced the lists. 
Replacing the subject `Mary' with `John' in the vignette, replacing the colon with a comma or randomly permuting the preface or the prompt strings did not affect the results (Fig.~\ref{fig:vignettes-ctrl}, right, Appendix \ref{sec:appendix}).
By contrast, the same perturbations reduced retrieval effects for Wikitext-103 (Fig.~\ref{fig:vignettes-ctrl}, left, Appendix \ref{sec:appendix}), supporting the conclusion that larger training corpus size contributes to robustness of transformer retrieval.

\paragraph{Transformer retrieval was robust to the number of items being retrieved.} \label{ex:set-size}

In studies of human short-term memory, performance degrades as the number of items that need to be retained increases (``set-size effects'', \citealt{oberauer_benchmarks_2018}). 
Is our LMs' short-term memory similarly taxed by increasing the set size?
We varied the number of tokens to be held in memory with $N^{\textit{tokens}} \in \{3, 5, 7, 10\}$.
For this comparison, the length of the intervening text was kept at 26 tokens.
Results reported in Fig.~\ref{fig:set-size} show that for GPT-2, verbatim recall was, for the most part, consistent across the different set sizes.
Repeat surprisal increased monotonically with set size only when the order of nouns in second list, either semantically coherent or arbitrary, was permuted.\footnote{This increase in surprisal with set size for permuted sequences is to be expected, of course, because, if the model has perfect memory of the list of tokens, but cannot predict the order in which they will reoccur, then its probability of guessing the next item in a permuted list where $k$ items have yet to be observed will be $1/k$, and the mean value of $k$ is larger for larger set sizes.}
For the smaller Wiktiext-103 transformer, repeat surprisal showed a slight increase with set size further indicating that retrieval robustness increases with training corpus size.

\begin{figure*}
\centering
\includegraphics[width=0.85\linewidth]{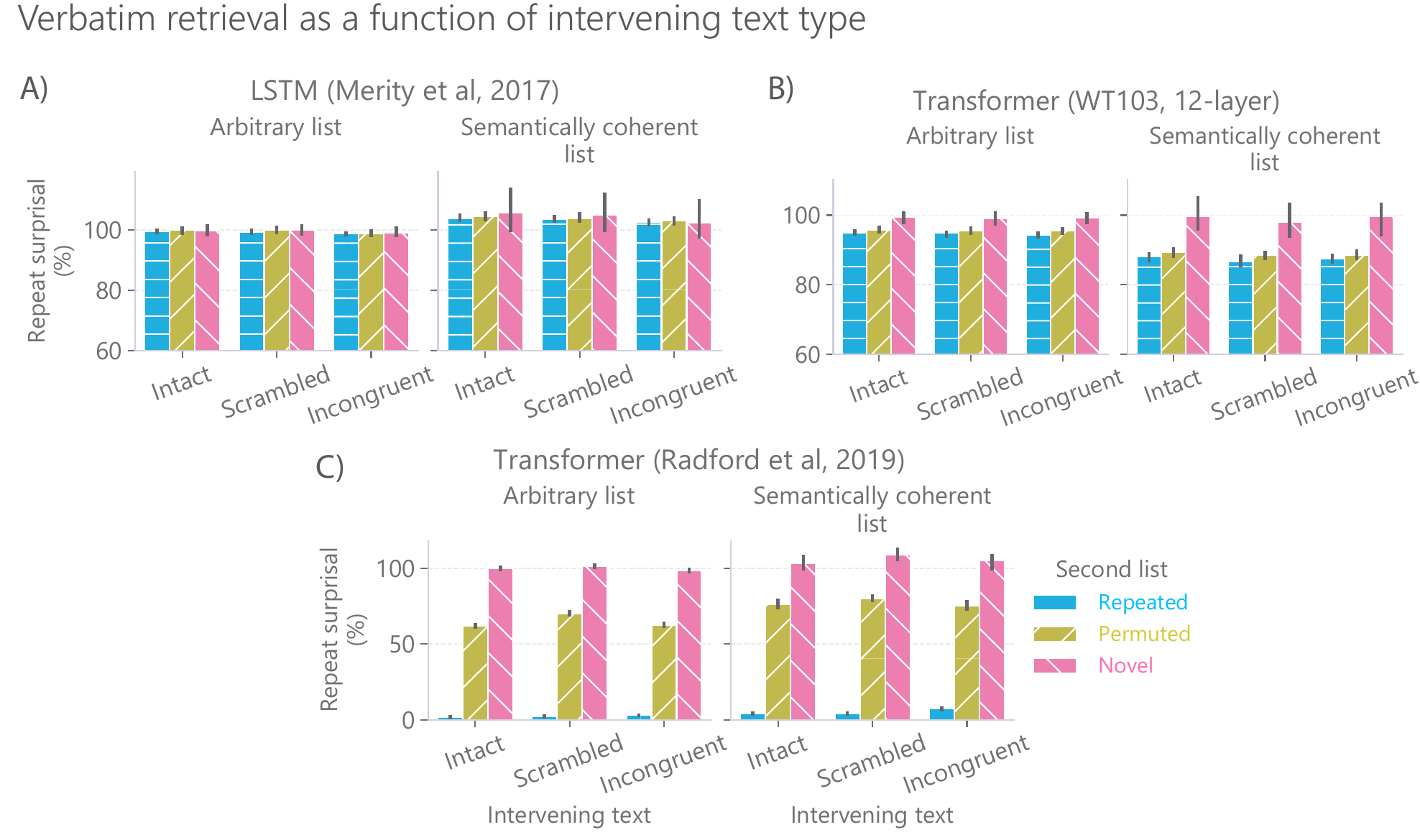}
\caption{LM memory retrieval for different intervening texts.
We plot relative list-averaged surprisal over all non-initial tokens in lists.
Points show group median (over $N^{\textit{list}}$ = 230). 
Error bars denote 95\% confidence interval around the median (bootstrap estimate).
Note that in the top-row plots y-axis starts at 60\%.}
\label{fig:inter-text-type}
\end{figure*}

\paragraph{Transformer retrieval was robust to the length and content of intervening text, but scrambling the intervening text reduced retrieval of order information.}
For how long are individual items retained in the memory of the LM?
We tested this by varying the length of the intervening text for $N^{\textit{tokens}} \in \{26, 99, 194, 435\}$ (see Fig.~\ref{fig:design}, panel~B).
To generate longer intervening text samples, we continued the narrative established by the initial preface string (``Before the meeting, Mary wrote down the following list of words:'').
All intervening text strings ended with the same prompt string (``When she got back, she read the list again:'') which introduced the second list.

Memory retrieval in GPT-2 was largely invariant to the size of the intervening text between the first and second lists (Fig.~\ref{fig:set-size}, B and C, respectively). The Wikitext-103 transformer exhibited small repeat surprisal increase over intervening text length, suggesting it's memory retrieval was less robust to maintenance over long distances compared to GPT-2.
All in all, the results suggest that the two transformers were retrieving prior nouns using a form of direct indexing of the relevant words from the input buffer, rather than implementing a generic memory heuristic, such as predicting that the nouns that have occurred in the most recent 20 tokens will recur.

Increasing the length of \textit{well-formed, semantically coherent} intervening text does not, then, interfere with memory retrieval in the transformer. 
In models of human memory, current context, such as immediately preceding text, can indeed be used as a cue for recalling the encoded items \citep[]{kahana_computational_2020}.
Does the transformers' capacity to retrieve copies of past nouns rely on the content and structure of the intervening text?
We tested this by creating incongruent and scrambled versions of the longest intervening text (435 tokens).
An incongruent condition was created by using intervening text that was syntactically well-formed but semantically incongruent with respect to the preface.
The scrambled version was created by randomly permuting the tokens of the intervening text.

The transformers' retrieval of past tokens was largely unaffected by the specific content of the intervening text, as long as the intervening text was coherent/well-formed (Fig.~\ref{fig:inter-text-type}).
However, in GPT-2, median surprisal across permuted arbitrary lists of nouns increased by $8\%$ when the intervening text was scrambled (Fig.~\ref{fig:inter-text-type}, bottom) compared to well-formed text.
This suggests that GPT-2 relied on narrative coherence of the intervening text, rather than its aggregate semantic content alone, as a cue for retrieving the ordering information of arbitrary word lists.

\paragraph{Transformer verbatim recall is learned, guided by attention, and requires increase in size.}

Having shown that the transformer LMs could flexibly and robustly retrieve words and their ordering verbatim from short-term memory (Figs. \ref{fig:set-size} and \ref{fig:inter-text-type}), we next asked: is this ability learned, or does it derive directly from the architecture? To address this question, we re-ran the experiment with varying number of tokens in lists with a randomly initialized transformer model (architecture as in Section \ref{sec:models}).
This random-weights model was unable to retrieve words or their order: for example, repeat surprisal remained at $100\%$ relative to first lists regardless of whether or not the nouns in the second list have appeared before (Fig.~\ref{fig:trf-ctrl}, top, Appendix \ref{sec:appendix}).

\begin{figure*}[ht]
\centering
\includegraphics[width=0.75\linewidth]{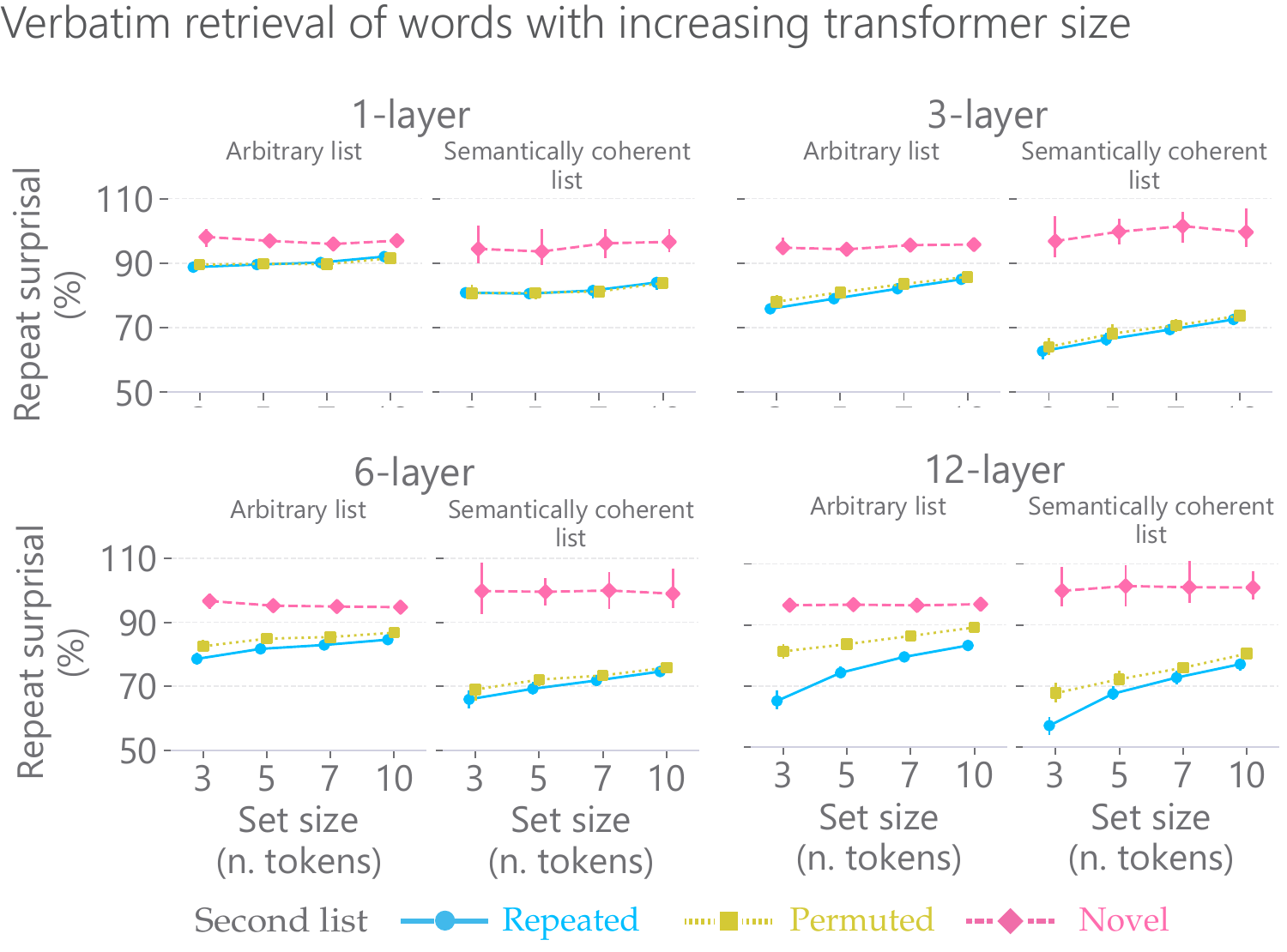}
\caption{LM memory retrieval for models of different sizes.
Reported is relative list-averaged surprisal over all non-initial tokens in lists.
Points show group median (over $N^{list}$ = 230). 
Error bars denote 95\% confidence interval around the median (bootstrap estimate).
Note that in these plots y-axis starts at 50\%.}
\label{fig:trf-layer}
\end{figure*}

Next we tested whether the transformers' ability to recall past tokens depended on the attention mechanism \citep[]{bahdanau_neural_2014, vaswani_attention_2017} which allows it, in principle, to use all past words, weighted according to their relevance, for next word prediction. 
To test for the role of attention in verbatim retrieval, we randomly permuted the rows of key and query matrices in each of the 12 attention layers of GPT-2 and reran the experiment with varying number of tokens in lists.
The shuffled-attention model retained some capacity to retrieve past nouns (Fig.~\ref{fig:trf-ctrl}, bottom, Appendix~\ref{sec:appendix}), but the effect was greatly reduced. For example, repeat surprisal for lists of $N=10$ semantically coherent nouns was at $90\%$ relative to first lists for shuffled-attention, compared with $3\%$ for the intact model.
Intriguingly, this shuffled-attention model showed the same surprisal for repeated and permuted lists, indicating that it was no longer accessing word order information from the original list. 
Thus, the attention mechanism is necessary for transformers to index past nouns and their order from memory.

Finally, a deep layered architecture is a key characteristic of transformers and performance typically scales with model size \citep[][]{radford_language_2019, kaplan_scaling_2020}.
Does the capacity to perform verbatim recall depend on model size?
To address this question, we trained transformers with 1, 3, 6 and 12 layers on the Wikitext-103 dataset.
Consistent with the hypothesis that size -- in addition to architecture -- is crucial, the smaller 1- and 3-layer models showed a modest verbatim recall capacity, but were not sensitive to order (e.g. the 3-layer model shows $85\%$ repeat surprisal for repeated and permuted lists of $N=10$ tokens, Fig.~\ref{fig:trf-layer}).
Sensitivity to order progressively emerged in 6- and 12-layer models, where in the 12-layer model repeat surprisal levels were $5\%$ lower for repeated relative to permuted 10-token lists (Fig.~\ref{fig:trf-layer}).
While this result confirms that even transformers trained on smaller amounts of text can exhibit short-term memory with sufficient increase in complexity, it remains unclear whether it is the increased depth or the parameter count alone that contribute to this increase in performance.

\section{LSTM Results}
\label{sec:lstm}

\paragraph{The LSTM retrieves gist-like memories over short intervening distances, facilitated by semantic coherence.} \label{sec:lstm_no_ordering}

The LSTM language model expected nouns in the second list to belong to the same semantic category as the first list, and especially to the category of the earliest nouns in the first list.
If the intervening text was no longer than 26 tokens, LSTM repeat surprisal across non-initial token positions (Fig.~\ref{fig:set-size}, A) showed a modest decrease ($5\%$) relative to first list, but only when the nouns in the first and second lists came from the same semantic category.
Examining surprisal values broken down by token position in the list (Fig.~\ref{fig:per-token}, top) shows that in semantically coherent lists of nouns, surprisal was higher for novel lists than for repeated or permuted lists, but this memory effect was only present for tokens near the beginning of the list. 

\begin{figure}
\centering
\includegraphics[width=1\linewidth]{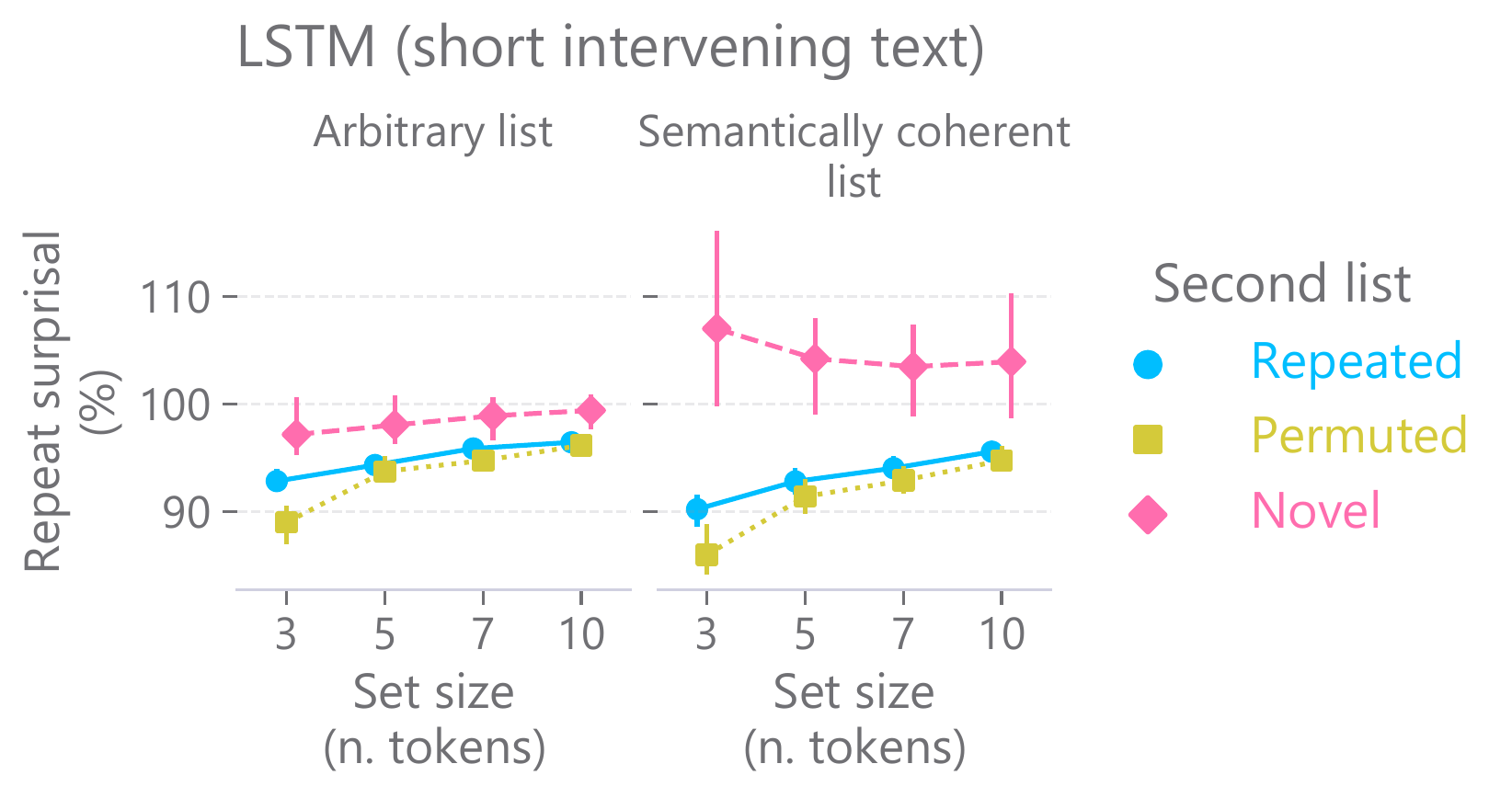}
\caption{LSTM verbatim token retrieval for varying number of tokens being retrieved at short (4-token) intervening text.
Reported is proportion of list-averaged surprisal on second relative to first list of nouns (repeat surprisal).
Points show group median (over $N^{list}$ = 230). 
Error bars denote 95\% confidence interval around the median (bootstrap estimate).}
\label{fig:set-size_filler-short_awd-lstm}
\end{figure}

In light of this limited evidence for retrieval in the LSTM across 26 intervening tokens, we examined whether the LSTM retrieves more successfully over shorter intervals. We reduced the intervening text to 4 tokens of coherent text (``Before the meeting, Mary wrote down the following lists of words. One was: <first list> \textbf{And the other:} <second list>''). 
In this short-range retrieval setting, we now observed a small reduction of relative repeat surprisal of $5\%$ and $4\%$ for arbitrary lists of 3 or 5 nouns, respectively, as well as a stronger reductions ranging from $12\%$ (3-token list) to $5\%$ (10-token list) for semantically coherent lists (Fig.~\ref{fig:set-size_filler-short_awd-lstm}).

Overall, the reduction in surprisal was comparable for repeated and permuted lists, indicating that the LSTM did not predict that words would occur in their original order.
Taken together, the experiments described in the section suggest that the LSTM retrieves a semantic gist of the prior list, rather than individual tokens in order.
Consistent with this notion of an aggregate semantic memory, we found that retrieval was stronger for semantically coherent lists, for which an aggregated semantic representation would be closer to each of the individual words in the list.

\section{Discussion}

Short-term memory---the capacity to temporarily store and access recent context for current processing---is a crucial component of language prediction. 
In this paper, we introduced a paradigm for characterizing a language model's short-term memory capabilities, based on retrieval of verbatim content (sequences of nouns) from prior context, and used this paradigm to analyze LMs with transformer and LSTM architectures.

The transformers we tested were able to access verbatim information -- individual tokens and their order -- from past context.
Furthermore, this verbatim retrieval was learned and largely \textit{resilient} to interference from intervening context.
This indicates that the models (especially those trained on the largest corpora) implemented, via learning, a high-resolution memory system.
The ability to access individual tokens may in turn support functions that rely on token indexing, akin to the functionality of the general-purpose working memory (WM) buffer proposed in cognitive science \citep{baddeley_working_2003}.

Such flexible WM could subserve the reported ability of transformers to rapidly generalize to new tasks at runtime \citep{brown_language_2020}, also known as ``in-context learning''.
Indeed, in concurrent work to ours, \citet{olsson_context_2022} observed that small (2 or 3-layer) attention-only transformers developed attention heads that functioned as so-called ``induction heads''. These effectively performed pattern matching by looking over the past context for any occurrences of the current token and predicting the same (or similar) sequence completions.
Attention heads that learned this basic inductive computation were also shown to perform more general in-context learning for complex tasks such as language translation.
Similarly, it has been suggested that in standard RNNs such meta-learning requires a short-term memory mechanism known as fast weights \citep{schmidhuber_learning_1992, ba_using_2016} which can be thought of as analogous to self-attention in transformers \citep{schlag_linear_2021}.

However, a highly resilient verbatim memory system could also be disadvantageous if it causes the LM to place too much confidence on verbatim features of prior context for next-word prediction. Indeed, text generated from a transformer LM's predictions can be highly repetitive \citep{holtzman_curious_2020} -- it is possible that an over-reliance on accessing short-term memory may underlie this tendency.

In contrast to the transformers, the LSTM model only retrieved a coarse semantic category of previous lists, without fine-grained information about word order, and was only able to do so when the intervening text was short. This is in spite of the fact that the LSTM had a larger parameter count than the transformer models and obtained comparable perplexity on WikiText103 (Table \ref{tab:model_architectures}).
The tendency of LSTMs to rely on the fuzzy representation of past context for next-word prediction has been reported previously \citep{khandelwal_sharp_2018}.
Whereas in sequence-to-sequence tasks requiring recall of short lists of pseudowords, recurrent neural networks are a good model of human short-term memory \citep{botvinick_short-term_2006}, later research has shown that the copying capacity of LSTMs does not generalize to longer sequences of symbols \citep{grefenstette_learning_2015}.

Is tracking a shallow representation of context always a limitation?
Not necessarily.
Humans frequently maintain a ``good-enough'' (i.e. gist-like) representation of context \citep{ferreira_good_2007}.
When the potential for memory capacity is limited (e.g. when context must be compressed to a single hidden state as in an RNN) maintaining a broad, gist-like -- as opposed to token-specific -- memory of context may be more \textit{efficient} overall.

The memory paradigm and the measure of repeat surprisal introduced here allowed us to pinpoint computational differences in how neural LMs put their architectural capacities to use for storing and accessing context in short-term memory when processing English text.
While our decision to use autoregressive (left-to-right) LMs was ultimately based on our initial cognitive psycholinguistic motivation, it may be fruitful to apply our paradigm to other classes of transformer models, for example, bidirectional encoder-only transformers such as BERT \citep{devlin_bert_2019} and encoder-decoder models such as T5 \citep{raffel_exploring_2020}. These architectures have gained traction in applied NLP settings and it would be informative to test whether this paradigm can provide diagnostic value for LM performance on other benchmarks. Similarly, if the compressed context representation in LSTMs serves as a short-term memory bottleneck, it would be instructive to test LSTM LM architectures when explicitly augmented with attention \citep{bahdanau_neural_2014} or a copy-mechanism \citep{gu_incorporating_2016}. Finally, our attention-ablation experiment in the transformer was performed uniformly across layers; future studies could focus on targeted ablations of specific attention heads to pinpoint the mechanistic locus of short-term memory \cite{olsson_context_2022}.

\section{Conclusions}

Pretrained language models, and self-supervised predictive learning broadly, have received increased attention in terms of their (in)sufficiency as a framework for achieving feats of human-like language processing \citep{kaplan_scaling_2020, linzen_syntactic_2021}.
Here, akin to the line of work evaluating cognitive linguistic capacities of neural LMs \citep{futrell_neural_2019, ritter_cognitive_2017}, we tested the ability of language models to perform an important aspect of human intelligence for natural language --- flexibly accessing items from short-term memory --- and showed that the transformer model, even though not trained with a short-term memory objective, retrieved remarkably detailed representations of past context. This capacity emerged from training: a transformer trained on a small amount of data showed more modest retrieval abilities. The retrieval abilities of LSTM LMs, by contrast, were different; the LSTM maintained a summary representation of the list, which was not sensitive to word order. We conclude that our paradigm can illuminate the memory systems that arise in neural language models.

\section{Broader Impact}

The research reported here addresses a specific, basic research question about the functional organization of short-term memory in contemporary language processing algorithms. Although from a broader perspective, the nature of (working) memory is likely an important question in developing human-like artificial intelligence systems deployed in real-life scenarios, it is, in our opinion, unlikely that the results reported here could pose or lead to novel societal risks as we are primarily trying to better the understanding of the already developed systems.

\section*{Acknowledgements}

The authors gratefully acknowledge the support of the National Institutes of Mental Health (grant R01MH119099). The research presented here also benefited from the discussions and feedback during the research visit (by KA) in the context of the collaborative grant ``Working Memory Based Assessment of Large Language Models'' at the Department of Language Technologies, Institute Jozef Stefan. The visit was in part financially supported by the Slovenian Research Agency (grant BI-US/22-24-170).
Finally, this work was supported in part through the NYU IT High Performance Computing resources, services, and staff expertise.

\bibliography{library}

\begin{thebibliography}{38}
\expandafter\ifx\csname natexlab\endcsname\relax\def\natexlab#1{#1}\fi

\bibitem[{Ba et~al.(2016)Ba, Hinton, Mnih, Leibo, and Ionescu}]{ba_using_2016}
Jimmy Ba, Geoffrey Hinton, Volodymyr Mnih, Joel~Z. Leibo, and Catalin Ionescu.
  2016.
\newblock Using fast weights to attend to the recent past.
\newblock In \emph{Proceedings of the 30th {International} {Conference} on
  {Neural} {Information} {Processing} {Systems}}, {NIPS}'16, pages 4338--4346,
  Red Hook, NY, USA. Curran Associates Inc.

\bibitem[{Baddeley(2003)}]{baddeley_working_2003}
Alan Baddeley. 2003.
\newblock \href {https://doi.org/10.1038/nrn1201} {Working memory: looking back
  and looking forward}.
\newblock \emph{Nature Reviews Neuroscience}, 4(10):829--839.

\bibitem[{Bahdanau et~al.(2014)Bahdanau, Cho, and
  Bengio}]{bahdanau_neural_2014}
Dzmitry Bahdanau, Kyunghyun Cho, and Yoshua Bengio. 2014.
\newblock \href {http://arxiv.org/abs/1409.0473} {Neural machine translation by
  jointly learning to align and translate}.
\newblock \emph{CoRR}, abs/1409.0473.

\bibitem[{Botvinick and Plaut(2006)}]{botvinick_short-term_2006}
Matthew~M. Botvinick and David~C. Plaut. 2006.
\newblock \href {https://doi.org/10.1037/0033-295X.113.2.201} {Short-term
  memory for serial order: {A} recurrent neural network model}.
\newblock \emph{Psychological Review}, 113(2):201--233.

\bibitem[{Brown et~al.(2020)Brown, Mann, Ryder, Subbiah, Kaplan, Dhariwal,
  Neelakantan, Shyam, Sastry, Askell, Agarwal, Herbert-Voss, Krueger, Henighan,
  Child, Ramesh, Ziegler, Wu, Winter, Hesse, Chen, Sigler, Litwin, Gray, Chess,
  Clark, Berner, McCandlish, Radford, Sutskever, and
  Amodei}]{brown_language_2020}
Tom Brown, Benjamin Mann, Nick Ryder, Melanie Subbiah, Jared~D Kaplan, Prafulla
  Dhariwal, Arvind Neelakantan, Pranav Shyam, Girish Sastry, Amanda Askell,
  Sandhini Agarwal, Ariel Herbert-Voss, Gretchen Krueger, Tom Henighan, Rewon
  Child, Aditya Ramesh, Daniel Ziegler, Jeffrey Wu, Clemens Winter, Chris
  Hesse, Mark Chen, Eric Sigler, Mateusz Litwin, Scott Gray, Benjamin Chess,
  Jack Clark, Christopher Berner, Sam McCandlish, Alec Radford, Ilya Sutskever,
  and Dario Amodei. 2020.
\newblock \href
  {https://proceedings.neurips.cc/paper/2020/file/1457c0d6bfcb4967418bfb8ac142f64a-Paper.pdf}
  {Language models are few-shot learners}.
\newblock In \emph{Advances in {Neural} {Information} {Processing} {Systems}},
  volume~33, pages 1877--1901. Curran Associates, Inc.

\bibitem[{Devlin et~al.(2019)Devlin, Chang, Lee, and
  Toutanova}]{devlin_bert_2019}
Jacob Devlin, Ming-Wei Chang, Kenton Lee, and Kristina Toutanova. 2019.
\newblock \href {https://doi.org/10.18653/v1/N19-1423} {{BERT}: {Pre}-training
  of deep bidirectional transformers for language understanding}.
\newblock In \emph{Proceedings of the 2019 {Conference} of the {North}
  {American} {Chapter} of the {Association} for {Computational} {Linguistics}:
  {Human} {Language} {Technologies}, {Volume} 1 ({Long} and {Short} {Papers})},
  pages 4171--4186, Minneapolis, Minnesota. Association for Computational
  Linguistics.

\bibitem[{Ferreira and Patson(2007)}]{ferreira_good_2007}
Fernanda Ferreira and Nikole~D. Patson. 2007.
\newblock \href {https://doi.org/10.1111/j.1749-818X.2007.00007.x} {The
  ‘{Good} {Enough}’ approach to language comprehension}.
\newblock \emph{Language and Linguistics Compass}, 1(1-2):71--83.

\bibitem[{Futrell et~al.(2018)Futrell, Wilcox, Morita, and
  Levy}]{futrell_rnns_2018}
Richard Futrell, Ethan Wilcox, Takashi Morita, and Roger Levy. 2018.
\newblock \href {http://arxiv.org/abs/1809.01329} {{RNNs} as psycholinguistic
  subjects: {Syntactic} state and grammatical dependency}.
\newblock \emph{arXiv:1809.01329 [cs]}.
\newblock ArXiv: 1809.01329.

\bibitem[{Futrell et~al.(2019)Futrell, Wilcox, Morita, Qian, Ballesteros, and
  Levy}]{futrell_neural_2019}
Richard Futrell, Ethan Wilcox, Takashi Morita, Peng Qian, Miguel Ballesteros,
  and Roger Levy. 2019.
\newblock \href {https://doi.org/10.18653/v1/N19-1004} {Neural language models
  as psycholinguistic subjects: {Representations} of syntactic state}.
\newblock In \emph{Proceedings of the 2019 {Conference} of the {North}
  {American} {Chapter} of the {Association} for {Computational} {Linguistics}:
  {Human} {Language} {Technologies}, {Volume} 1 ({Long} and {Short} {Papers})},
  pages 32--42, Minneapolis, Minnesota. Association for Computational
  Linguistics.

\bibitem[{Grefenstette et~al.(2015)Grefenstette, Hermann, Suleyman, and
  Blunsom}]{grefenstette_learning_2015}
Edward Grefenstette, Karl~Moritz Hermann, Mustafa Suleyman, and Phil Blunsom.
  2015.
\newblock Learning to transduce with unbounded memory.
\newblock In \emph{Proceedings of the 28th {International} {Conference} on
  {Neural} {Information} {Processing} {Systems} - {Volume} 2}, {NIPS}'15, pages
  1828--1836, Cambridge, MA, USA. MIT Press.

\bibitem[{Gu et~al.(2016)Gu, Lu, Li, and Li}]{gu_incorporating_2016}
Jiatao Gu, Zhengdong Lu, Hang Li, and Victor~O.K. Li. 2016.
\newblock \href {https://doi.org/10.18653/v1/P16-1154} {Incorporating copying
  mechanism in sequence-to-sequence learning}.
\newblock In \emph{Proceedings of the 54th {Annual} {Meeting} of the
  {Association} for {Computational} {Linguistics} ({Volume} 1: {Long}
  {Papers})}, pages 1631--1640, Berlin, Germany. Association for Computational
  Linguistics.

\bibitem[{Gulordava et~al.(2018)Gulordava, Bojanowski, Grave, Linzen, and
  Baroni}]{gulordava_colorless_2018}
Kristina Gulordava, Piotr Bojanowski, Edouard Grave, Tal Linzen, and Marco
  Baroni. 2018.
\newblock \href {https://doi.org/10.18653/v1/N18-1108} {Colorless green
  recurrent networks dream hierarchically}.
\newblock In \emph{Proceedings of the 2018 {Conference} of the {North}
  {American} {Chapter} of the {Association} for {Computational} {Linguistics}:
  {Human} {Language} {Technologies}, {Volume} 1 ({Long} {Papers})}, pages
  1195--1205, New Orleans, Louisiana. Association for Computational
  Linguistics.

\bibitem[{Hochreiter and Schmidhuber(1997)}]{hochreiter_long_1997}
Sepp Hochreiter and Jürgen Schmidhuber. 1997.
\newblock \href {https://doi.org/10.1162/neco.1997.9.8.1735} {Long short-term
  memory}.
\newblock \emph{Neural Computation}, 9(8):1735--1780.

\bibitem[{Holtzman et~al.(2020)Holtzman, Buys, Du, Forbes, and
  Choi}]{holtzman_curious_2020}
Ari Holtzman, Jan Buys, Li~Du, Maxwell Forbes, and Yejin Choi. 2020.
\newblock \href {https://iclr.cc/virtual_2020/poster_rygGQyrFvH.html} {The
  curious case of neural text degeneration}.

\bibitem[{Kahana(2020)}]{kahana_computational_2020}
Michael~J. Kahana. 2020.
\newblock \href {https://doi.org/10.1146/annurev-psych-010418-103358}
  {Computational models of memory search}.
\newblock \emph{Annual Review of Psychology}, 71(1):107--138.

\bibitem[{Kaplan et~al.(2020)Kaplan, McCandlish, Henighan, Brown, Chess, Child,
  Gray, Radford, Wu, and Amodei}]{kaplan_scaling_2020}
Jared Kaplan, Sam McCandlish, Tom Henighan, Tom~B. Brown, Benjamin Chess, Rewon
  Child, Scott Gray, Alec Radford, Jeffrey Wu, and Dario Amodei. 2020.
\newblock \href {http://arxiv.org/abs/2001.08361} {Scaling laws for neural
  language models}.
\newblock \emph{arXiv:2001.08361 [cs, stat]}.
\newblock ArXiv: 2001.08361.

\bibitem[{Khandelwal et~al.(2018)Khandelwal, He, Qi, and
  Jurafsky}]{khandelwal_sharp_2018}
Urvashi Khandelwal, He~He, Peng Qi, and Dan Jurafsky. 2018.
\newblock \href {https://doi.org/10.18653/v1/P18-1027} {Sharp nearby, fuzzy far
  away: {How} neural language models use context}.
\newblock In \emph{Proceedings of the 56th {Annual} {Meeting} of the
  {Association} for {Computational} {Linguistics} ({Volume} 1: {Long}
  {Papers})}, pages 284--294, Melbourne, Australia. Association for
  Computational Linguistics.

\bibitem[{Lakretz et~al.(2021)Lakretz, Hupkes, Vergallito, Marelli, Baroni, and
  Dehaene}]{lakretz_mechanisms_2021}
Yair Lakretz, Dieuwke Hupkes, Alessandra Vergallito, Marco Marelli, Marco
  Baroni, and Stanislas Dehaene. 2021.
\newblock \href {https://doi.org/10.1016/j.cognition.2021.104699} {Mechanisms
  for handling nested dependencies in neural-network language models and
  humans}.
\newblock \emph{Cognition}, 213:104699.

\bibitem[{Lakretz et~al.(2019)Lakretz, Kruszewski, Desbordes, Hupkes, Dehaene,
  and Baroni}]{lakretz_emergence_2019}
Yair Lakretz, German Kruszewski, Theo Desbordes, Dieuwke Hupkes, Stanislas
  Dehaene, and Marco Baroni. 2019.
\newblock \href {https://doi.org/10.18653/v1/N19-1002} {The emergence of number
  and syntax units in {LSTM} language models}.
\newblock In \emph{Proceedings of the 2019 {Conference} of the {North}
  {American} {Chapter} of the {Association} for {Computational} {Linguistics}:
  {Human} {Language} {Technologies}, {Volume} 1 ({Long} and {Short} {Papers})},
  pages 11--20, Minneapolis, Minnesota. Association for Computational
  Linguistics.

\bibitem[{Linzen and Baroni(2021)}]{linzen_syntactic_2021}
Tal Linzen and Marco Baroni. 2021.
\newblock \href {https://doi.org/10.1146/annurev-linguistics-032020-051035}
  {Syntactic structure from deep learning}.
\newblock \emph{Annual Review of Linguistics}, 7(1):195--212.
\newblock \_eprint: https://doi.org/10.1146/annurev-linguistics-032020-051035.

\bibitem[{Linzen et~al.(2016)Linzen, Dupoux, and
  Goldberg}]{linzen_assessing_2016}
Tal Linzen, Emmanuel Dupoux, and Yoav Goldberg. 2016.
\newblock \href {https://doi.org/10.1162/tacl_a_00115} {Assessing the ability
  of {LSTMs} to learn {Syntax}-sensitive dependencies}.
\newblock \emph{Transactions of the Association for Computational Linguistics},
  4:521--535.

\bibitem[{Marvin and Linzen(2018)}]{marvin_targeted_2018}
Rebecca Marvin and Tal Linzen. 2018.
\newblock \href {https://doi.org/10.18653/v1/D18-1151} {Targeted syntactic
  evaluation of language models}.
\newblock In \emph{Proceedings of the 2018 {Conference} on {Empirical}
  {Methods} in {Natural} {Language} {Processing}}, pages 1192--1202, Brussels,
  Belgium. Association for Computational Linguistics.

\bibitem[{Merity et~al.(2018)Merity, Keskar, and
  Socher}]{merity_regularizing_2018}
Stephen Merity, Nitish~Shirish Keskar, and Richard Socher. 2018.
\newblock \href {https://openreview.net/forum?id=SyyGPP0TZ} {Regularizing and
  optimizing {LSTM} language models}.

\bibitem[{Merity et~al.(2016)Merity, Xiong, Bradbury, and
  Socher}]{merity_pointer_2016}
Stephen Merity, Caiming Xiong, James Bradbury, and Richard Socher. 2016.
\newblock \href {https://openreview.net/forum?id=Byj72udxe} {Pointer sentinel
  mixture models}.

\bibitem[{Oberauer et~al.(2018)Oberauer, Lewandowsky, Awh, Brown, Conway,
  Cowan, Donkin, Farrell, Hitch, Hurlstone, Ma, Morey, Nee, Schweppe, Vergauwe,
  and Ward}]{oberauer_benchmarks_2018}
Klaus Oberauer, Stephan Lewandowsky, Edward Awh, Gordon D.~A. Brown, Andrew
  Conway, Nelson Cowan, Christopher Donkin, Simon Farrell, Graham~J. Hitch,
  Mark~J. Hurlstone, Wei~Ji Ma, Candice~C. Morey, Derek~Evan Nee, Judith
  Schweppe, Evie Vergauwe, and Geoff Ward. 2018.
\newblock \href {https://doi.org/10.1037/bul0000153} {Benchmarks for models of
  short-term and working memory.}
\newblock \emph{Psychological Bulletin}, 144(9):885--958.

\bibitem[{O'Connor and Andreas(2021)}]{oconnor_what_2021}
Joe O'Connor and Jacob Andreas. 2021.
\newblock \href {https://doi.org/10.18653/v1/2021.acl-long.70} {What context
  features can transformer language models use?}
\newblock In \emph{Proceedings of the 59th {Annual} {Meeting} of the
  {Association} for {Computational} {Linguistics} and the 11th {International}
  {Joint} {Conference} on {Natural} {Language} {Processing} ({Volume} 1: {Long}
  {Papers})}, pages 851--864, Online. Association for Computational
  Linguistics.

\bibitem[{Olsson et~al.(2022)Olsson, Elhage, Nanda, Joseph, DasSarma, Henighan,
  Mann, Askell, Bai, Chen, Conerly, Drain, Ganguli, Hatfield-Dodds, Hernandez,
  Johnston, Jones, Kernion, Lovitt, Ndousse, Amodei, Brown, Clark, Kaplan,
  McCandlish, and Olah}]{olsson_context_2022}
Catherine Olsson, Nelson Elhage, Neel Nanda, Nicholas Joseph, Nova DasSarma,
  Tom Henighan, Ben Mann, Amanda Askell, Yuntao Bai, Anna Chen, Tom Conerly,
  Dawn Drain, Deep Ganguli, Zac Hatfield-Dodds, Danny Hernandez, Scott
  Johnston, Andy Jones, Jackson Kernion, Liane Lovitt, Kamal Ndousse, Dario
  Amodei, Tom Brown, Jack Clark, Jared Kaplan, Sam McCandlish, and Chris Olah.
  2022.
\newblock In-context learning and induction heads.
\newblock \emph{Transformer Circuits Thread}.

\bibitem[{Radford et~al.(2019)Radford, Wu, Child, Luan, Amodei, and
  Sutskever}]{radford_language_2019}
Alec Radford, Jeff Wu, Rewon Child, David Luan, Dario Amodei, and Ilya
  Sutskever. 2019.
\newblock \href
  {https://d4mucfpksywv.cloudfront.net/better-language-models/language-models.pdf}
  {Language models are unsupervised multitask learners}.

\bibitem[{Raffel et~al.(2020)Raffel, Shazeer, Roberts, Lee, Narang, Matena,
  Zhou, Li, and Liu}]{raffel_exploring_2020}
Colin Raffel, Noam Shazeer, Adam Roberts, Katherine Lee, Sharan Narang, Michael
  Matena, Yanqi Zhou, Wei Li, and Peter~J. Liu. 2020.
\newblock Exploring the limits of transfer learning with a unified text-to-text
  transformer.
\newblock \emph{J. Mach. Learn. Res.}, 21(1).
\newblock Publisher: JMLR.org.

\bibitem[{Ritter et~al.(2017)Ritter, Barrett, Santoro, and
  Botvinick}]{ritter_cognitive_2017}
Samuel Ritter, David G.~T. Barrett, Adam Santoro, and Matt~M. Botvinick. 2017.
\newblock \href {https://proceedings.mlr.press/v70/ritter17a.html} {Cognitive
  psychology for deep neural networks: {A} shape bias case study}.
\newblock In \emph{Proceedings of the 34th {International} {Conference} on
  {Machine} {Learning}}, pages 2940--2949. PMLR.
\newblock ISSN: 2640-3498.

\bibitem[{Schlag et~al.(2021)Schlag, Irie, and
  Schmidhuber}]{schlag_linear_2021}
Imanol Schlag, Kazuki Irie, and Jürgen Schmidhuber. 2021.
\newblock \href {https://proceedings.mlr.press/v139/schlag21a.html} {Linear
  transformers are secretly fast weight programmers}.
\newblock In \emph{Proceedings of the 38th {International} {Conference} on
  {Machine} {Learning}}, volume 139 of \emph{Proceedings of {Machine}
  {Learning} {Research}}, pages 9355--9366. PMLR.

\bibitem[{Schmidhuber(1992)}]{schmidhuber_learning_1992}
Jürgen Schmidhuber. 1992.
\newblock \href {https://doi.org/10.1162/neco.1992.4.1.131} {Learning to
  control fast-weight memories: {An} alternative to dynamic recurrent
  networks}.
\newblock \emph{Neural Computation}, 4(1):131--139.

\bibitem[{Subramanian et~al.(2020)Subramanian, Collobert, Ranzato, and
  Boureau}]{subramanian_multi-scale_2020}
Sandeep Subramanian, Ronan Collobert, Marc'Aurelio Ranzato, and Y.-Lan Boureau.
  2020.
\newblock \href {http://arxiv.org/abs/2005.00581} {Multi-scale transformer
  language models}.
\newblock \emph{arXiv:2005.00581 [cs]}.
\newblock ArXiv: 2005.00581.

\bibitem[{Tay et~al.(2020)Tay, Dehghani, Bahri, and
  Metzler}]{tay_efficient_2020}
Yi~Tay, Mostafa Dehghani, Dara Bahri, and Donald Metzler. 2020.
\newblock \href {http://arxiv.org/abs/2009.06732} {Efficient transformers: {A}
  survey}.
\newblock \emph{arXiv:2009.06732 [cs]}.
\newblock ArXiv: 2009.06732 version: 2.

\bibitem[{Vaswani et~al.(2017)Vaswani, Shazeer, Parmar, Uszkoreit, Jones,
  Gomez, Kaiser, and Polosukhin}]{vaswani_attention_2017}
Ashish Vaswani, Noam Shazeer, Niki Parmar, Jakob Uszkoreit, Llion Jones,
  Aidan~N. Gomez, Łukasz Kaiser, and Illia Polosukhin. 2017.
\newblock Attention is all you need.
\newblock In \emph{Proceedings of the 31st {International} {Conference} on
  {Neural} {Information} {Processing} {Systems}}, {NIPS}'17, pages 6000--6010,
  Red Hook, NY, USA. Curran Associates Inc.

\bibitem[{Weston et~al.(2015)Weston, Chopra, and Bordes}]{weston_memory_2015}
Jason Weston, Sumit Chopra, and Antoine Bordes. 2015.
\newblock Memory networks.
\newblock In \emph{3rd {International} {Conference} on {Learning}
  {Representations}, {ICLR} 2015}, San Diego. 3rd International Conference on
  Learning Representations.

\bibitem[{Wolf et~al.(2020)Wolf, Debut, Sanh, Chaumond, Delangue, Moi, Cistac,
  Rault, Louf, Funtowicz, Davison, Shleifer, von Platen, Ma, Jernite, Plu, Xu,
  Le~Scao, Gugger, Drame, Lhoest, and Rush}]{wolf_transformers_2020}
Thomas Wolf, Lysandre Debut, Victor Sanh, Julien Chaumond, Clement Delangue,
  Anthony Moi, Pierric Cistac, Tim Rault, Remi Louf, Morgan Funtowicz, Joe
  Davison, Sam Shleifer, Patrick von Platen, Clara Ma, Yacine Jernite, Julien
  Plu, Canwen Xu, Teven Le~Scao, Sylvain Gugger, Mariama Drame, Quentin Lhoest,
  and Alexander Rush. 2020.
\newblock \href {https://doi.org/10.18653/v1/2020.emnlp-demos.6} {Transformers:
  {State}-of-the-{Art} natural language processing}.
\newblock In \emph{Proceedings of the 2020 {Conference} on {Empirical}
  {Methods} in {Natural} {Language} {Processing}: {System} {Demonstrations}},
  pages 38--45, Online. Association for Computational Linguistics.

\bibitem[{Yogatama et~al.(2018)Yogatama, Miao, Melis, Ling, Kuncoro, Dyer, and
  Blunsom}]{yogatama_memory_2018}
Dani Yogatama, Yishu Miao, Gabor Melis, Wang Ling, Adhiguna Kuncoro, Chris
  Dyer, and Phil Blunsom. 2018.
\newblock \href {https://openreview.net/forum?id=SkFqf0lAZ} {Memory
  architectures in recurrent neural network language models}.

\end{thebibliography}
\bibliographystyle{acl_natbib}

\appendix

\section{Appendix}
\label{sec:appendix}

\subsection{Vignettes}\label{sec:vignettes}

\onecolumn
\textbf{Intact intervening text:}

Before the meeting, Mary wrote down the following list of words:

$W_{1}, W_{2}, ..., W_{N}$

$intervening\_text_1$: After the meeting, she took a break and had a cup of coffee. When she got back, she read the list again: $W_{1}, W_{2}, ..., W_{N}$

$intervening\_text_2$: After the meeting, Mary went for a walk. It was a busy day and she needed a break. Outside was really beautiful and warm and the flowers in the park were blooming. When she got back, she read the list again: $W_{1}, W_{2}, ..., W_{N}$

$intervening\_text_3$: After the meeting, Mary went for a walk. It was a busy day and she needed a break. Outside was really beautiful and warm and the flowers in the park were blooming. While she was walking, she listened to the wonderful bird songs. During the walk, Mary could not stop thinking about the meeting. She was thinking about the discussions she had with her coworkers. Luckily, she met her neighbors Sarah and Ryan and they talked briefly. When she got back, she read the list again: $W_{1}, W_{2}, ..., W_{N}$

$intervening\_text_4$: After the meeting, Mary went for a walk. It was a busy day and she needed a break. Outside was really beautiful and warm and the flowers in the park were blooming. While she was walking, she listened to the wonderful bird songs. During the walk, Mary could not stop thinking about the meeting. She was thinking about the discussions she had with her coworkers. Luckily, she met her neighbors Sarah and Ryan and they talked briefly. The couple has just moved to the area from a different city. Mary thought they were very a lovely couple and made good company. They were just getting to know the neighborhood and this was their first time in the park. Mary was curious what were their first impressions of the town. The neighborhood felt very safe to them and they absolutely loved the park. This was only their second time visiting the park. There was so much to discover, so many winding paths and hidden gardens. When she got back, she read the list again: $W_{1}, W_{2}, ..., W_{N}$

$intervening\_text_5$: After the meeting, Mary went for a walk. It was a busy day and she needed a break. Outside was really beautiful and warm and the flowers in the park were blooming. While she was walking, she listened to the wonderful bird songs. During the walk, Mary could not stop thinking about the meeting. She was thinking about the discussions she had with her coworkers. Luckily, she met her neighbors Sarah and Ryan and they talked briefly. The couple has just moved to the area from a different city. Mary thought they were very a lovely couple and made good company. They were just getting to know the neighborhood and this was their first time in the park. Mary was curious what were their first impressions of the town. The neighborhood felt very safe to them and they absolutely loved the park. This was only their second time visiting the park. There was so much to discover, so many winding paths and hidden gardens. It was not a big park by any means, but it offered a quiet refuge where one can escape the worries of everyday life. It also offered opportunities to do sports of all kinds. Young people from around the area played basketball, football, or volleyball. Others took part in outdoor workout sessions. Young families were going on a stroll with their children. Finally, there were so many people who brought their dogs for a walk. It was incredibly satisfying to see the joy our animal friends get when you throw them a ball. All this diversity of people and activities made a walk in this park a truly rewarding and relaxing daily routine. In fact, Sarah and Ryan were thinking of getting a dog. They have not fully decided yet but they really wanted to spend more time outdoors. Mary liked dogs as well, but she was more of a cat person herself. She and her husband had two cats. One was two and the other four years old. They were very independent and spent most of their time outdoors. Mary thought having an animal was a great idea. They talked for a little bit and then Sarah and Ryan invited her to come over for a cup of coffee. Mary said she had time over the weekend. When she got back, she read the list again: $W_{1}, W_{2}, ..., W_{N}$

\newpage
\onecolumn
\textbf{Scrambled intervening text:}

Before the meeting, Mary wrote down the following list of words:

$W_{1}, W_{2}, ..., W_{N}$

$intervening\_text_1$: After a break, a cup and coffee of had she the took meeting. When she got back, she read the list again: $W_{1}, W_{2}, ..., W_{N}$

$intervening\_text_2$: Outside the the beautiful and park flowers blooming were in and was warm really. After, walk for Mary the a went meeting. It needed busy break she day was a and a. When she got back, she read the list again: $W_{1}, W_{2}, ..., W_{N}$

$intervening\_text_3$: Luckily and and met Sarah they Ryan briefly talked her, neighbors she. Thinking during, stop meeting the not about Mary the could walk. The while walking to songs bird listened wonderful, she she was. After, walk for Mary the a went meeting. Had she about she coworkers her with the was discussions thinking. Outside the the beautiful and park flowers blooming were in and was warm really. It needed busy break she day was a and a. When she got back, she read the list again: $W_{1}, W_{2}, ..., W_{N}$

$intervening\_text_4$: First they their was neighborhood getting and the in park the this to were time know just. There paths so much, and many gardens hidden winding to was discover so. The while walking to songs bird listened wonderful, she she was. Had she about she coworkers her with the was discussions thinking. From the just area city different the a moved couple to has. The absolutely and very them loved they park the safe neighborhood to felt. Outside the the beautiful and park flowers blooming were in and was warm really. And Mary were couple company good lovely made very thought a they. Luckily and and met Sarah they Ryan briefly talked her, neighbors she. Thinking during, stop meeting the not about Mary the could walk. After, walk for Mary the a went meeting. Their this park visiting second was the time only. Impressions Mary what first town the of were was their curious. It needed busy break she day was a and a. When she got back, she read the list again: $W_{1}, W_{2}, ..., W_{N}$

$intervening\_text_5$: It needed busy break she day was a and a. First they their was neighborhood getting and the in park the this to were time know just. Had she about she coworkers her with the was discussions thinking. Of they independent most outdoors time their and were spent very. Get it friends them our joy satisfying when the throw ball a animal to was you see incredibly. The while walking to songs bird listened wonderful, she she was. Weekend had time Mary said the over she. An Mary idea a animal thought great was having. Mary a she was as but cat of herself person more well liked, dogs. It of opportunities kinds sports to also all do offered. Cats husband had she and two her. They spend they really fully but more to outdoors time wanted decided have yet not. A a and of rewarding park all in made this activities relaxing routine daily truly walk people this and diversity. There paths so much, and many gardens hidden winding to was discover so. Finally dogs who were people for brought walk a their so, many there. Luckily and and met Sarah they Ryan briefly talked her, neighbors she. The absolutely and very them loved they park the safe neighborhood to felt. Outside the the beautiful and park flowers blooming were in and was warm really. Young football basketball around played,, volleyball or the people area from. Their this park visiting second was the time only. To Sarah a a for her they Ryan then invited and cup coffee of over come and little talked bit for. From the just area city different the a moved couple to has. And Mary were couple company good lovely made very thought a they. Young with going children stroll on families their a were. Worries a means escape where a offered but one refuge can it by any it the quiet of life everyday, big was park not. Of in Sarah thinking dog a were getting and fact Ryan,. Thinking during, stop meeting the not about Mary the could walk. After, walk for Mary the a went meeting. And one old four the was years other two. Impressions Mary what first town the of were was their curious. Sessions in outdoor others took workout part. When she got back, she read the list again: $W_{1}, W_{2}, ..., W_{N}$

\newpage
\onecolumn
\textbf{Incongruent intervening text:}

Before the meeting, Mary wrote down the following list of words:

$W_{1}, W_{2}, ..., W_{N}$

$intervening\_text_1$: There is a voice in the waters of the great sea. It calls to man continually. When she got back, she read the list again: $W_{1}, W_{2}, ..., W_{N}$

$intervening\_text_2$: Sometimes it thunders in the tempest, when the waves leap high and strong and the wild winds shriek and roar. Sometimes it whispers in the calm, small voice, as if to solicit our regard. When she got back, she read the list again: $W_{1}, W_{2}, ..., W_{N}$

$intervening\_text_3$: After the meeting, Mary went for a walk. It was a busy day and she needed a break. Outside was really beautiful and warm and the flowers in the park were blooming. The sea has much to say; far more than could possibly be comprehended in one volume, however large. It tells us of the doings of man on its broad bosom, from the day in which he first ventured to paddle along shore to the day when he launched his great iron ship, and rushed out to sea. When she got back, she read the list again: $W_{1}, W_{2}, ..., W_{N}$

$intervening\_text_4$: After the meeting, Mary went for a walk. It was a busy day and she needed a break. Outside was really beautiful and warm and the flowers in the park were blooming. The sea has much to say; far more than could possibly be comprehended in one volume, however large. It tells us of the doings of man on its broad bosom, from the day in which he first ventured to paddle along shore to the day when he launched his great iron ship, and rushed out to sea. Before proceeding to the consideration of the wonders connected with and contained in the sea, we shall treat of the composition of the sea itself and of its extent, depth, and bottom. What is the sea made of? Salt water, is the ready reply that rises naturally to every lip. But to this we add the question, what is salt water? To these queries we give the following reply, which, we doubt not, will rather surprise some of our readers. The salt of the ocean varies considerably in different parts. When she got back, she read the list again: $W_{1}, W_{2}, ..., W_{N}$

$intervening\_text_5$: After the meeting, Mary went for a walk. It was a busy day and she needed a break. Outside was really beautiful and warm and the flowers in the park were blooming. The sea has much to say; far more than could possibly be comprehended in one volume, however large. It tells us of the doings of man on its broad bosom, from the day in which he first ventured to paddle along shore to the day when he launched his great iron ship, and rushed out to sea. Before proceeding to the consideration of the wonders connected with and contained in the sea, we shall treat of the composition of the sea itself and of its extent, depth, and bottom. What is the sea made of? Salt water, is the ready reply that rises naturally to every lip. But to this we add the question, what is salt water? To these queries we give the following reply, which, we doubt not, will rather surprise some of our readers. The salt of the ocean varies considerably in different parts. Near the equator, the great heat carries up a larger proportion of water by evaporation than in the more temperate regions. Thus, as salt is not removed by evaporation, the ocean in the torrid zone is salter than in the temperate or frigid zones. The salts of the sea, and other substances contained in it, are conveyed there by the fresh water streams that pour into it from all the continent of the world Here, as these substances cannot be evaporated, they would accumulate to such a degree as to render the ocean uninhabitable by living creatures.The operations of the ocean are manifold. But we cannot speak of these things without making passing reference to the operations of water, as that wonder-working agent of which the ocean constitutes but a part. Nothing in this world is ever lost or annihilated. As the ocean receives all the water that flows from the land, so it returns that water, fresh and pure, in the shape of vapour, to the skies. where, in the form of clouds, it is conveyed to those parts of the earth where its presence is most needed. After having gladdened the heart of man by driving his mills and causing his food to grow, it finds its way again into the sea: and thus the good work goes on with ceaseless regularity. When she got back, she read the list again: $W_{1}, W_{2}, ..., W_{N}$\footnote{The incongruent intervening text was sampled from: ``The ocean and its wonder'' by R. M. Ballantyne (obtained from: \url{https://www.gutenberg.org/ebooks/21754}).}

\newpage
\onecolumn
\textbf{Short intervening text:}

Before the meeting, Mary wrote down the following lists of words. One was:

$W_{1}, W_{2}, ..., W_{N}$

$intervening\_text_1$: And the other: $W_{1}, W_{2}, ..., W_{N}$

\twocolumn

\subsection{Noun Lists}

\small
\begin{table*}
\centering
\caption{Arbitrary lists of nouns used in present experiments.}
\label{tab:arbitrary_nouns}
\small
\begin{tabular}{ll}
\toprule
{} &                                                                                    list \\
\midrule
1  &             patience, notion, movie, women, canoe, novel, folly, silver, eagle, center. \\
2  &     pleasure, pattern, leader, culture, worker, master, meadow, writer, apple, costume. \\
3  &          paper, belief, factor, total, comrade, angle, battle, pistol, nothing, riches. \\
4  &       cabin, doorway, candle, parent, monarch, kindness, lover, copy, soldier, kingdom. \\
5  &   future, legend, problem, flavor, prairie, forehead, illness, planet, canvas, chamber. \\
6  &        oven, patient, daughter, bubble, colour, product, echo, pepper, fountain, music. \\
7  &    village, shipping, beauty, football, merit, autumn, lumber, research, resort, rival. \\
8  &    county, muscle, vapor, shepherd, sickness, herald, value, mission, finger, building. \\
9  &        iron, onion, opera, attack, prison, butter, interest, colonel, commerce, beggar. \\
10 &  blanket, marriage, ticket, baby, treasure, event, weakness, cottage, cotton, judgment. \\
11 &     summer, bottom, meaning, campaign, voyage, cannon, helmet, thunder, hatred, stanza. \\
12 &      effort, province, parcel, temple, river, major, meeting, career, bargain, chimney. \\
13 &        acre, fortune, motive, question, service, minute, tiger, author, sorrow, parlor. \\
14 &       motor, lawyer, powder, habit, mountain, district, learning, leather, hero, water. \\
15 &         orange, letter, acid, stocking, olive, garden, feeling, motion, compass, model. \\
16 &      island, theory, person, season, supper, reason, patent, picture, custom, twilight. \\
17 &      dragon, pillow, aspect, chairman, marble, horror, justice, danger, bedroom, canal. \\
18 &  writing, pocket, training, circuit, cousin, chapter, quarter, button, turkey, surface. \\
19 &       sailor, matter, darkness, scatter, captain, tunnel, method, wagon, effect, arrow. \\
20 &       image, butcher, anchor, scholar, compound, tribute, victim, lily, witness, widow. \\
21 &         candy, window, detail, ocean, program, traffic, feather, array, pilot, silence. \\
22 &       vessel, robber, banner, kitten, lemon, failure, princess, painter, bullet, rifle. \\
23 &            engine, timber, harbour, party, level, money, single, system, unit, traitor. \\
\bottomrule
\end{tabular}
\end{table*}

\small
\begin{table*}
\centering
\caption{Semantically coherent lists of nouns used in present experiments.}
\label{tab:semantic_nouns}
\small
\begin{tabular}{ll}
\toprule
{} &                                                                                                      list \\
\midrule
1  &                                   window, door, roof, wall, floor, ceiling, room, basement, hearth, hall. \\
2  &                                                 leg, arms, head, eye, foot, nose, finger, ear, hand, toe. \\
3  &         sailboat, destroyer, battleship, cruiser, submarine, yacht, canoe, freighter, tugboat, steamship. \\
4  &                            robin, sparrow, heron, eagle, crow, hawk, parrot, pigeon, woodpecker, vulture. \\
5  &                              apple, pear, banana, peach, grape, cherry, plum, grapefruit, lemon, apricot. \\
6  &                                   hammer, saw, nails, level, plane, chisel, ruler, wrench, drill, screws. \\
7  &                             hurricane, tornado, rain, snow, hail, storm, wind, cyclone, clouds, sunshine. \\
8  &                oxygen, hydrogen, nitrogen, carbon, sodium, sulphur, helium, chlorine, calcium, potassium. \\
9  &  chemistry, physics, psychology, biology, zoology, botany, astronomy, mathematics, geology, microbiology. \\
10 &                         piano, drum, trumpet, violin, clarinet, flute, guitar, saxophone, trombone, oboe. \\
11 &                                              knife, spoon, fork, pan, pot, stove, bowl, mixer, cup, dish. \\
12 &                                    trout, shark, herring, perch, salmon, tuna, goldfish, cod, carp, pike. \\
13 &              football, baseball, basketball, tennis, swimming, soccer, golf, hockey, lacrosse, badminton. \\
14 &          doctor, lawyer, teacher, dentist, engineer, professor, carpenter, salesman, nurse, psychologist. \\
15 &                                      oak, maple, pine, elm, birch, spruce, redwood, walnut, fir, hickory. \\
16 &                                     shirt, socks, pants, shoes, blouse, skirt, coat, dress, hat, sweater. \\
17 &    cancer, measles, tuberculosis, polio, malaria, leukemia, pneumonia, smallpox, influenza, encephalitis. \\
18 &                                   mountain, hill, valley, river, rock, lake, canyon, tundra, ocean, cave. \\
19 &                     murder, rape, robbery, theft, assault, arson, kidnapping, larceny, adultery, battery. \\
20 &                                            log, cat, horse, cow, lion, tiger, elephant, pig, bear, mouse. \\
21 &                                     fly, ant, bee, mosquito, spider, beetle, wasp, moth, flea, butterfly. \\
22 &                                      blue, red, green, yellow, black, purple, white, pink, brown, blonde. \\
23 &                                     cotton, wool, silk, rayon, linen, satin, velvet, denim, canvas, felt. \\
\bottomrule
\end{tabular}
\end{table*}

\subsection{Model Parameter Comparison}

Comparison of model parameters across the three main models used in the present study is reported in Table \ref{tab:model_architectures}.

\begin{table*}[!ht]
\centering
\caption{Comparison of main architectural and training parameters between models used in the current study.}
\label{tab:model_architectures}
\begin{tabular}{llll}
\toprule
\textbf{Model} &                GPT-2 & transformer (WT-103) &            AWD LSTM \\
\midrule
\textbf{Reference                 } &  Radford et al (2019) &                 ours &  Merity et al (2017) \\
\textbf{Nr. layers                } &                   12 &                   12 &                   3 \\
\textbf{Train set size            } &    40 (GB text data) &        40 (M tokens) &      102 (M tokens) \\
\textbf{Nr. parameters (M)        } &                  117 &                107.7 &                 182 \\
\textbf{Embedding size            } &                  768 &                  768 &                 400 \\
\textbf{Hidden size               } &                  768 &                  768 &               1,840 \\
\textbf{Vocabulary size           } &               50,257 &               28,439 &             267,735 \\
\textbf{Context window (n. tokens)} &                 1024 &                 1024 &                   - \\
\textbf{WikiText103 perplexity           } &                37.50 &                 40.3 &                41.9 \\
\bottomrule
\end{tabular}
\end{table*}

\subsection{LSTM Training Details}\label{sec:lstm_training_details}

The AWD LSTM model was trained using our own version of the original repository. The hyperparameters used for training are reported in Table \ref{tab:lstm_training_params} (essentially input arguments to the original training script which we used: \url{https://github.com/salesforce/awd-lstm-lm/blob/master/main.py}).

To deploy the training job on an HPC cluster, we used a single GPU (NVIDIA RTX8000), requested 14GB of RAM and a job time of 48 hours. This was sufficient for the model to converge to the perplexity reported in Table \ref{tab:model_architectures}.

\begin{table*}[!ht]
\centering
\caption{Hyperparameter setup for training AWD LSTM.}
\label{tab:lstm_training_params}
\begin{tabular}{ll}
\toprule
{} & Parameter value \\
\midrule
\textbf{Vocabulary size (nr. tokens)} &          267,735 \\
\textbf{Nr. layers                  } &               3 \\
\textbf{Input embedding size        } &             400 \\
\textbf{Hidden size                 } &            1840 \\
\textbf{Output dropout              } &             0.4 \\
\textbf{Embedding dropout           } &               0 \\
\textbf{Hidden dropout              } &            0.01 \\
\textbf{Input dropout               } &            0.01 \\
\textbf{Weight drop                 } &             0.2 \\
\textbf{Weight decay                } &         $1.2^{-6}$ \\
\textbf{Tie weights                 } &            True \\
\textbf{Learning rate               } &          $1^{-3}$ \\
\textbf{Epochs                      } &              44 \\
\textbf{Lr reduction (epochs)       } &        [25, 35] \\
\textbf{Batch size                  } &             128 \\
\textbf{Adam alpha                  } &               0 \\
\textbf{Adam beta                   } &               0 \\
\textbf{BPTT                        } &             200 \\
\bottomrule
\end{tabular}
\end{table*}

\subsection{Transformer Training Details}\label{sec:transformer_training_details}

Transformer training hyperparameters are reported in Table \ref{tab:model-train-info}. These are effectively input arguments to the HuggingFace \texttt{Trainer()} (\url{https://huggingface.co/transformers/v4.6.0/main_classes/trainer.html}) and \texttt{GPT2Config()} (\url{https://huggingface.co/transformers/v4.6.0/model_doc/gpt2.html#gpt2config}) classes.
The model was trained until convergence and training was stopped (early stopping) when the loss did not decrease for at least 0.01 bits in 5 consecutive evaluations.

To train the transformer model on a HPC cluster, we requested a single GPU (NVIDIA RTX8000) with 44GB RAM and 12 hours of job time.

\begin{table*}
\centering
\caption{Hyperparameters for the transformers trained as part of this work.}
\label{tab:model-train-info}
\begin{tabular}{lllll}
\toprule
{} &   1 layer &   3 layer &   6 layer &  12 layer \\
\midrule
\textbf{Activation function        } &  gelu\_new &  gelu\_new &  gelu\_new &  gelu\_new \\
\textbf{Nr. layers                    } &         1 &         3 &         6 &        12 \\
\textbf{Nr. heads                     } &         3 &         3 &         6 &        12 \\
\textbf{Context size                      } &      1024 &      1024 &      1024 &      1024 \\
\textbf{Causal mask dimensionality                } &      1024 &      1024 &      1024 &      1024 \\
\textbf{Vocabulary size                 } &     28,439 &     28,439 &     28,439 &     28,439 \\
\textbf{Per device train batch size} &        12 &        12 &        12 &        12 \\
\textbf{Per device eval batch size } &        12 &        12 &        12 &        12 \\
\textbf{Learning rate              } &   0.00007 &   0.00007 &   0.00007 &   0.00006 \\
\textbf{Adam beta1                 } &       0.6 &       0.6 &       0.6 &       0.6 \\
\textbf{Adam beta2                 } &      0.05 &      0.05 &      0.05 &       0.1 \\
\textbf{Nr. parameters (millions)               } &      29.7 &      43.9 &      65.2 &     107.7 \\
\bottomrule
\end{tabular}
\end{table*}

\subsection{Compute Resources for Short-term Memory Evaluation Tasks}

For a single job (single experimental condition, e.g., evaluating GPT-2 on vignettes with $N = 230$ input sequences containing exactly repeated, abstract noun lists of length 10 and intervening text set to 26 tokens), a single GPU device was used and we typically requested $\sim$12 hours of core-walltime and $\sim$ 4 GB of RAM. To evaluate the RNN models, requesting 06:00 (hh:mm) of walltime and 4GB was typically more than sufficient to avoid any memory overflows.

\begin{figure*}[ht]
\centering
\includegraphics[width=1\linewidth]{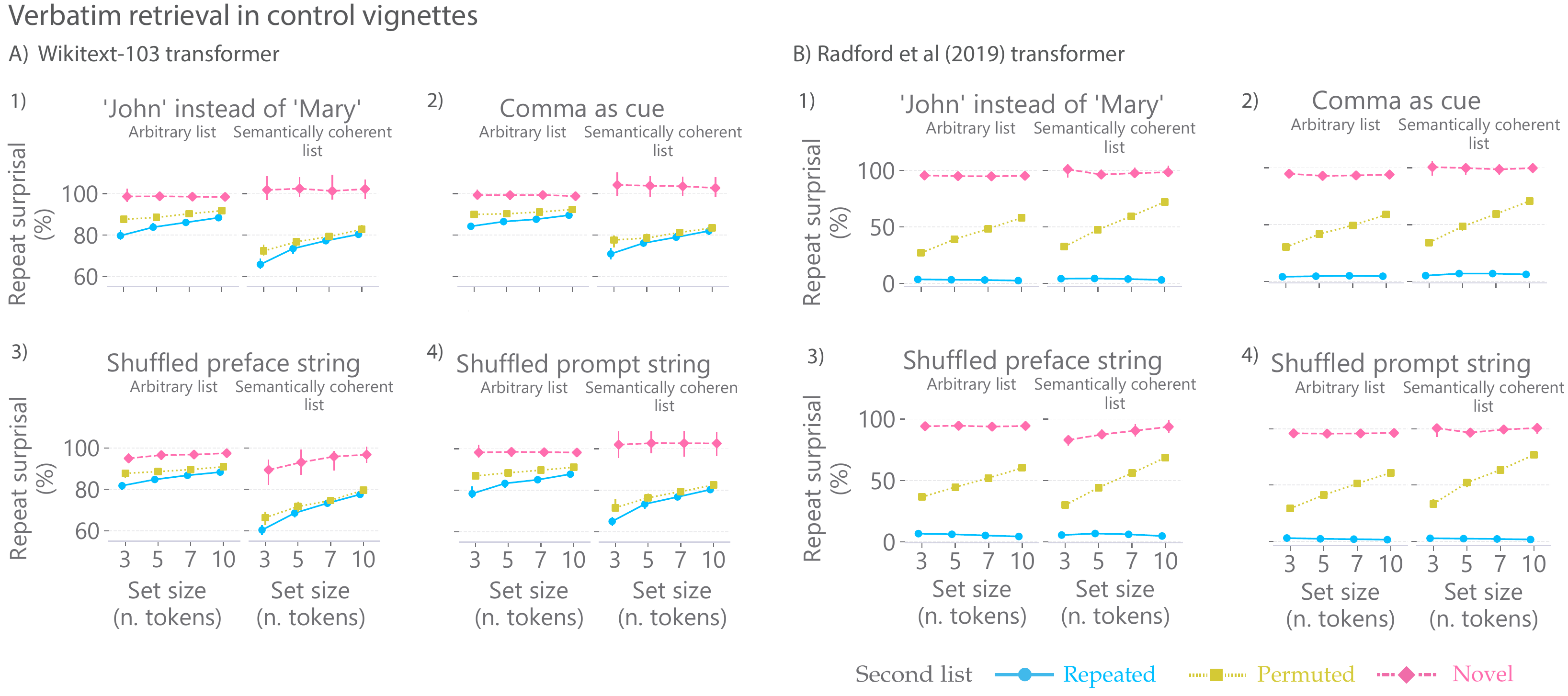}
\caption{Transformer memory retrieval results for control vignettes.
We report relative list-averaged surprisal over all non-initial tokens in lists (group median over $N^{list}$ = 230). 
Error bars denote 95\% confidence interval around the median (bootstrap estimate).
Note that in the Wikitext-103 plots the y-axis starts at 55\%.}
\label{fig:vignettes-ctrl}
\end{figure*}

\begin{figure*}
\centering
\includegraphics[width=0.6\linewidth]{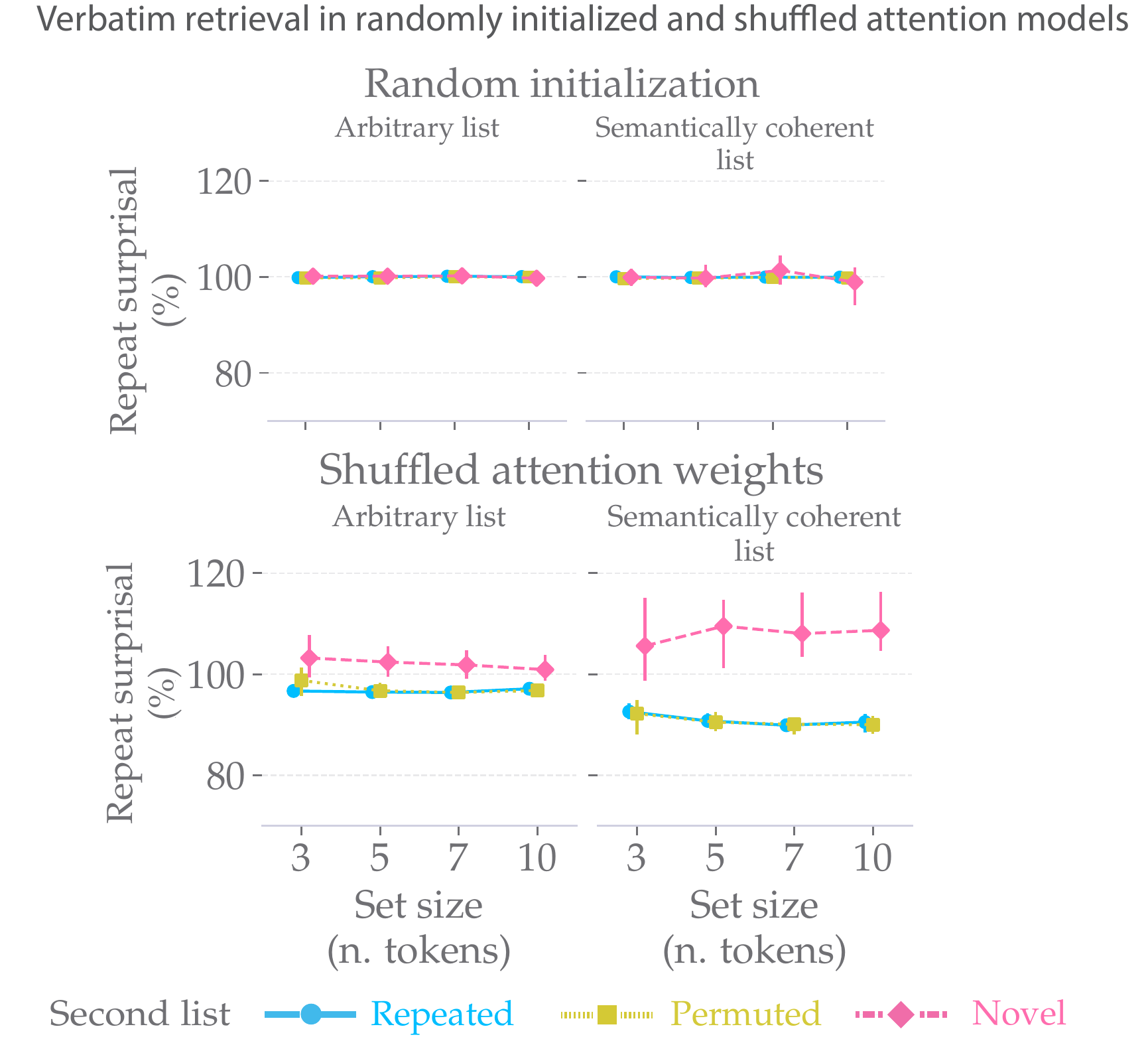}
\caption{Repeat surprisal for randomly initialized transformer LM and a transformer with permuted attention weights.
Reported is relative list-averaged surprisal over all non-initial tokens in lists only.
Points show group median (over $N^{list} = 230$). 
Error bars denote 95\% confidence interval around the median (bootstrap estimate).
Note that in these plots y-axis starts at 70\%.}
\label{fig:trf-ctrl}
\end{figure*}

\end{document}